\documentclass[times,singlecolumn,final]{elsarticle} %这边改变单列还是双列
\usepackage{medima}
\usepackage{framed,multirow}
\usepackage{amsmath,amssymb,amsfonts}
\usepackage{latexsym}

\usepackage{url}
\usepackage{xcolor}
\usepackage{float}

\usepackage{hyperref}

\usepackage{lineno}
\usepackage{bm}
\usepackage{color}
\usepackage{booktabs}
\usepackage{threeparttable}
\definecolor{newcolor}{rgb}{.8,.349,.1}

\journal{Medical Image Analysis}

\begin{document}
%\linenumbers  %加行号
\verso{K.Wang \textit{et~al.}}

\begin{frontmatter}

\title{TSdetector: Temporal-Spatial Self-correction Collaborative Learning for Colonoscopy Video Detection}%
%%\tnotetext[tnote1]{This is an example for title footnote coding.}

\author[1,2]{Kai-Ni Wang \fnref{fn1}}
%%\fntext[fn1]{This is author footnote for second author.}
\author[1,2]{Haolin Wang }
%% Third author's email
%%\ead{author3@author.com}
\author[1,2]{Guang-Quan Zhou \corref{cor1}}
\author[5]{Yangang Wang}
\author[6]{Ling Yang}
\author[3,4]{Yang Chen}
\author[7]{Shuo Li}
\cortext[cor1]{Corresponding author:
	e-mail address: guangquan.zhou@seu.edu.cn (Guang-Quan Zhou)}

\address[1]{ School of Biological Science and Medical Engineering, Southeast University, Nanjing, China}
\address[2]{Jiangsu Key Laboratory of Biomaterials and Devices, Southeast University, Nanjing, China}
\address[3]{Laboratory of Image Science and Technology, Southeast University, Nanjing, China}
\address[4]{Key Laboratory of Computer Network and Information Integration, Southeast University, Nanjing, China}
\address[5]{TUGE Healthcare, Nanjing, China}
\address[6]{Institute of Medical Technology, Peking University Health Science Center, China}
\address[7]{Department of Computer and Data Science and Department of Biomedical Engineering, Case Western Reserve University, USA}

\received{1 May 2013}
\finalform{10 May 2013}
\accepted{13 May 2013}
\availableonline{15 May 2013}
\communicated{S. Sarkar}

\begin{abstract}
%%%
CNN-based object detection models that strike a balance between performance and speed have been gradually used in polyp detection tasks. Nevertheless, accurately locating polyps within complex colonoscopy video scenes remains challenging since existing methods ignore two key issues: intra-sequence distribution heterogeneity and precision-confidence discrepancy. To address these challenges, we propose a novel Temporal-Spatial self-correction detector (TSdetector), which first integrates temporal-level consistency learning and spatial-level reliability learning to detect objects continuously. Technically, we first propose a global temporal-aware convolution, assembling the preceding information to dynamically guide the current convolution kernel to focus on global features between sequences. In addition, we designed a hierarchical queue integration mechanism to combine multi-temporal features through a progressive accumulation manner, fully leveraging contextual consistency information together with retaining long-sequence-dependency features. Meanwhile, at the spatial level, we advance a position-aware clustering to explore the spatial relationships among candidate boxes for recalibrating prediction confidence adaptively, thus eliminating redundant bounding boxes efficiently. The experimental results on three publicly available polyp video dataset show that TSdetector achieves the highest polyp detection rate and outperforms other state-of-the-art methods. The code can be available at https://github.com/soleilssss/TSdetector.

\end{abstract}


\end{frontmatter}

%\linenumbers

%% main text
\begin{figure*}[t]
	\centering
	\includegraphics[width=0.8\textwidth]{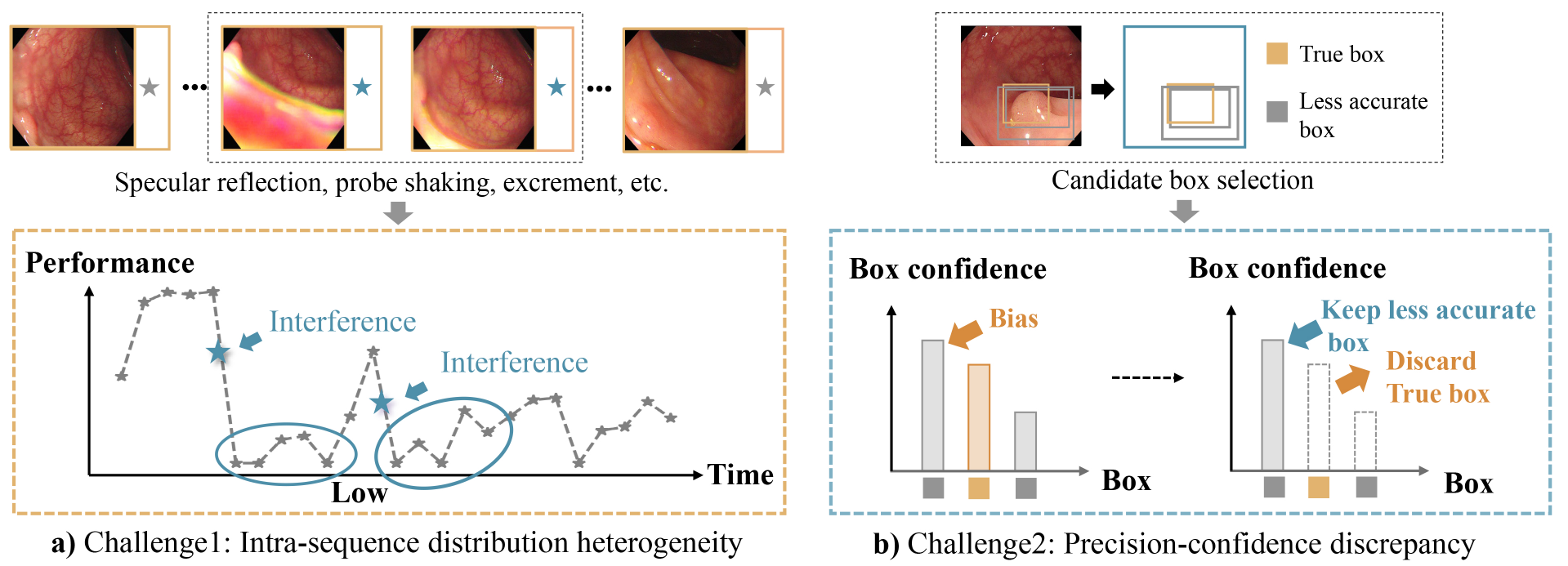}
	\caption{Two key challenges in video polyp detection: intra-sequence distribution heterogeneity and precision-confidence discrepancy.}
	\label{img1}
\end{figure*}

\section{Introduction}
\label{sec1}

CNN-based methods have become prevalent in object detection and have been deployed in the medical task of polyp detection \cite{bernal2017comparative, mamonov2014automated,jiang2023yona}. Generally speaking, two-stage detectors attain superior accuracy, whereas one-stage detectors can achieve a better trade-off between accuracy and performance \cite{yang2022real}. In fact, object detection models trained from high-quality images often fail to achieve satisfactory results when confronted with colonoscopy video scenarios \cite{zhang2020asynchronous}. A key question remains: How can we leverage a novel paradigm to compensate for the limitations of traditional CNN detection models?

Recently, many works have shown improvements in video polyp detection models \cite{puyal2022polyp,liu2022source, wang2022afp}, but two persistent challenges remain. 1) \textbf{Intra-sequence distribution heterogeneity} (Fig. \ref{img1} a). This refers to the diversity in the distribution of features within a sequence of frames in a video, specifically the differences observed between consecutive frames due to the dynamic nature of colonoscopy procedures. For instance, one frame may exhibit clear imagery, while the next may contain distortions or occlusions due to the movement of the probe or other factors. In the endoscopic video, intra-sequence distribution heterogeneity describes not only fluctuations in image quality, such as those caused by motion artifacts and specular reflections. Additionally, it encompasses changes in the appearance of objects, structures, or backgrounds within the frames due to factors like variations in brightness, angle changes, liquid interference, and instrument occlusion\cite{wang2023adaptive}. Instrument occlusion refers to the situation where the view of the endoscopic camera is blocked or partially blocked by the medical equipment. This distribution heterogeneity can pose a significant uncertainty for detection algorithms, as the varying image characteristics can distract the attention of the network \cite{ling2023probabilistic, wang2022ffcnet} and cause it to focus on irrelevant regions, leading to tracking failures. 2) \textbf{Precision-confidence discrepancy} (Fig. \ref{img1} b).  This issue arises when the bounding box with the highest confidence value may not necessarily be the true positive with the largest overlap with the ground truth box \cite{zheng2021enhancing}. Since models often select the candidate boxes with the highest confidence scores. This bias can lead to missing the most reliable proposals, as other objects with slightly lower scores are simply discarded.

\begin{figure*}[t]
	\centering
	\includegraphics[width=0.5\textwidth]{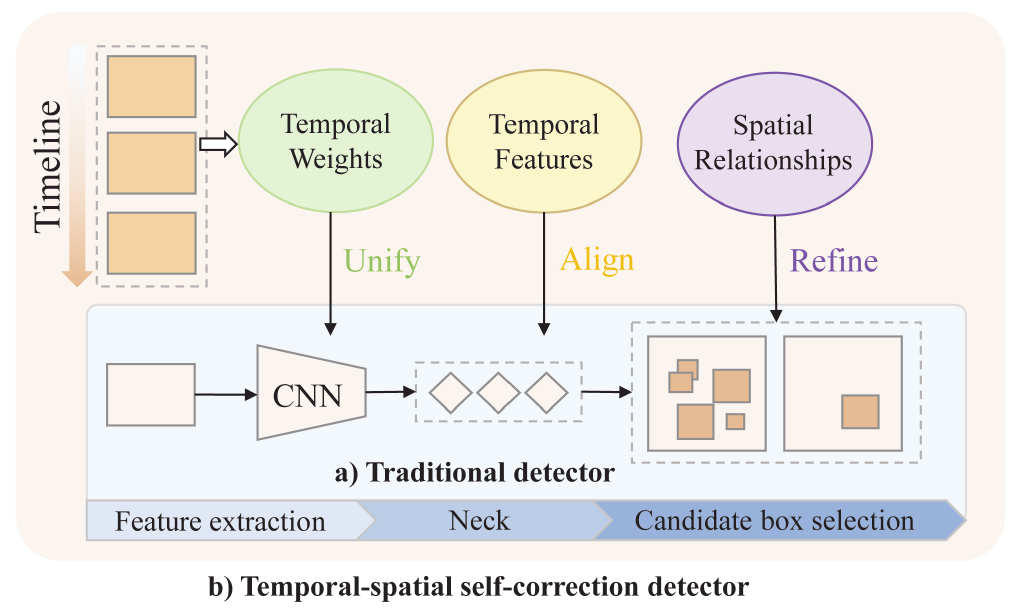}
	\caption{Comparison between classical detection model paradigms a) and our temporal-spatial self-correcting detector b). In contrast, our TSdetector utilizes spatial and temporal information to compensate for the limitations of traditional detection models from three perspectives.}
	\label{img2}
\end{figure*}

One problem is that many existing object detectors are designed to process each input frame or image independently, overlooking the valuable temporal cues in continuous video streams. Although prior methodologies, ranging from early approaches using traditional shape and texture models \cite{tajbakhsh2015automated} to recent attempts using convolutional neural networks \cite{qadir2019improving} or transformers, have demonstrated impressive performance \cite{tamhane2022colonoscopy}, there remains a performance gap when extending these methods to video-based polyp detection. This gap arises due to the additional temporal dimension in videos absent in single-frame images. Consequently, exploring the correlation and complementarity of nearby frames becomes crucial to compensate for possible image perturbations or model errors in a single image. Some works are dedicated to leveraging temporal context through one-shot aggregation of features and temporally deformable transformer networks \cite{wu2021multi}. However, high memory consumption and complex structure greatly hinder inference performance under real conditions.

Another problem is the current post-processing candidate box selection strategy, which completely relies on the confidence score output by the model, causing bottlenecks in target misses and positioning deviations. Positioning deviations denote potential inaccuracies in localizing detected polyps. As for end-to-end detectors, Non-Maximum Suppression (NMS) \cite{neubeck2006efficient} remains the most efficacious post-processing step for further enhancing accuracy and reducing inference time overhead. In contrast to the conventional NMS approach, Soft-NMS \cite{bodla2017soft} offers a more accommodating strategy by assigning reduced confidence values to bounding boxes instead of outright elimination, rendering it more suitable for scenarios involving occlusions. However, these NMS variants \cite{pathiraja2023multiclass} depend on the ranking of confidence scores, which may not always align with the true positive of the overlap ratio of the ground truth box (Fig. \ref{img1} b).  This inconsistency diminishes the reliability of confidence scores for obtaining optimal detection boxes. Intuitively, one potential solution is to calibrate the confidence scores of candidate boxes to be more reliable with performance.

\begin{figure}[!t]
	\centering
	\includegraphics[width=\textwidth]{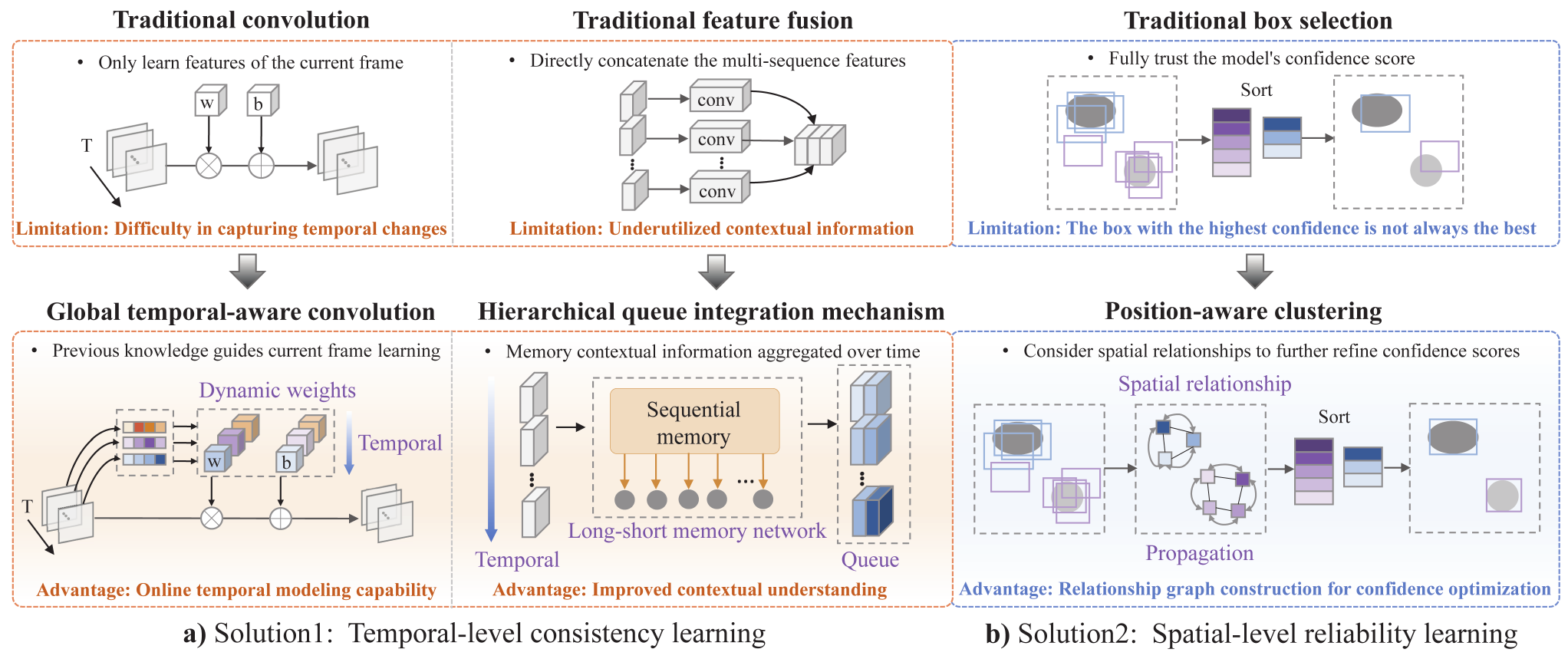}
	\caption{Comparison between the limitations of existing detection frameworks and the advantages of the proposed method. a) and b) represent two solution ideas for the challenge: temporal-level consistency learning and spatial-level reliability learning.}
	\label{img3}
\end{figure} 

To address the challenges above, we introduce a novel Temporal-Spatial self-correction network, dubbed TSdetector, for video polyp detection, which consists of two self-correction stages: temporal-level consistency learning and spatial-level reliability learning (Fig. \ref{img2}). 1) In the temporal-level consistency learning stage (Fig. \ref{img3} a),  we aim to guide feature extraction and fusion through temporal knowledge, thereby generating more refined proposals. We propose Global Temporal-aware Convolution (GT-Conv) whose convolution kernel weights are no longer static; instead, they are dynamically generated based on temporal contextual features. This dynamic adaptation complements the temporal modeling capabilities of conventional convolutions, further optimizing feature encoding. Additionally, we introduce the Hierarchical Queue Integration Mechanism (HQIM), a long short-term memory network that enables the capture of multi-temporal features in a progressive accumulation manner. HQIM memorizes and propagates previous information to the current frame, enhancing feature correlation to adapt to evolving data. 2) In the spatial-level reliability learning stage (Fig. \ref{img3} b), we aim to mitigate discrepancies between the confidence scores and the actual positive probabilities of candidate bounding boxes. We present the Position-Aware Clustering (PAC), a candidate box selection method grounded in spatial clustering. PAC leverages the relationships among candidate boxes to provide more comprehensive view-adaptive confidence. It effectively suppresses redundant boxes, thereby retaining the candidate boxes with the highest degree of overlap with the real boxes and reducing the risk of false positives. To summarize, our contributions are as follows: 

\begin{enumerate}
	\item We propose an innovative temporal-spatial self-correction network  for polyp video detection, leveraging both temporal-level and spatial-level optimization to compensate for CNN-based detection models.
	\item We design an effective global temporal-aware convolution and hierarchical queue integration mechanism, which mutually cooperate to integrate temporal information into the feature extraction and neck stages of the detector to cope with intra-sequence distribution heterogeneity.
	\item We present position-aware clustering, a new approach that leverages the relationship between candidate boxes to provide a more comprehensive view and adaptively adjusts the confidence, thereby improving the alignment between predictions and ground truth values.
	\item  Extensive experiments are conducted to verify the effectiveness of our method. TSdetector outperforms other existing methods and achieves the state-of-the-art results on three public polyp video datasets: SUN, CVC-ClinicDB, and PICCOLO.
\end{enumerate}

The rest of this paper is organized as follows. The next section \ref{sec2} reviews the related works. A detailed explanation of our proposed method is described in section \ref{sec3}. Sections \ref{sec4} and \ref{sec5} present experimental results and corresponding analysis. Finally, section \ref{sec6} concludes the proposed work.

\section{Related Works}
\label{sec2}
\subsection{Colonoscopy-related datasets}
%Automated polyp detection has been the subject of extensive research in recent years due to its potential to aid in diagnosis. Table \ref{tab1} provides a count of 14 current polyp detection datasets, which are divided into image-based (upper part) and video-based (lower part) based on the data format. 

\textbf{Image-based datasets.} Image-based datasets are usually used for polyp segmentation tasks due to mask-level annotations. The dense labeling masks required are labor-intensive and nearly impossible to fully label. Consequently, the labeling strategy is generally sampling, using single-frame labeling from consecutive video frames, resulting in small-scale image-based datasets. However, clinical colonoscopy is a continuous video task, and employing image-based methods directly on videos often leads to performance gaps.

\textbf{Video-based datasets.} Annotations for video-based datasets are typically in the form of bounding boxes and are widely utilized in object detection research. Except for SAU-Mayo \cite{tajbakhsh2015automated}, which is a dense label mask, its label is still a sampling type with only 3856 cases. Notably, the SUN dataset is the largest fully labeled dataset, making it a promising candidate for real-time polyp detection. Regrettably, there have been limited studies conducted on this dataset. Recently, Ji et al. introduced the SUN-SEG dataset \cite{ji2022video,ji2021progressively}, a multi-scale dataset that extends SUN by providing additional labels such as attributes, object masks, boundaries, graffiti, and polygons. This work provides a benchmark for video polyp segmentation and is a valuable resource for further research.

\subsection{Colonoscopy-related detection methods}
\textbf{Image-based methods.} In the early stages of image-based research, models heavily relied on feature extraction and selection. For instance, Tajbakhsh \cite{tajbakhsh2015automated} and Ameling \cite{ameling2009texture} utilized shape and texture features, respectively, for detection. Nonetheless, these approaches depend highly on handcrafted heuristics to assign appropriate feature representations, resulting in limited performance. With the advent of convolutional neural networks (CNNs), recent studies have shifted their focus to using deep learning models to automatically extract features. For instance, Mohammed et al. introduced Y-Net \cite{mohammed2018net}, comprising two encoders and one decoder to improve detection accuracy. Moreover, some studies choose to add auxiliary constraints on the original architecture \cite{itoh2022positive}, adding uncertainty estimation of categories and introducing weighted object activation maps.

\textbf{Video-based methods.} Since the image-based frame lacks information between frames, making it difficult to perceive the dynamic changes of objects, some works are devoted to exploring temporal features in continuous video frames. Qadir et al. leveraged temporal dependencies to improve the false positives of CNNs in colonoscopy videos \cite{qadir2019improving}. Ma et al. proposed a novel sample selection strategy \cite{ma2020polyp} that considers the temporal consistency of test videos. Xu et al. used structural similarity to measure the similarity between video frames to assist in making final decisions. Wu et al. proposed an efficient multi-frame collaborative framework \cite{wu2021multi}, spatio-temporal feature transformation. Overall, the above studies delve into temporal information between frames, exploring consistency, similarity, and feature fusion aspects. 

\subsection{Object detection methods}
Object detection methods are broadly divided into two-stage and one-stage detectors. Two-stage detectors \cite{he2017mask} are region proposal-based methods that first generate regions of interest from images and then classify candidate boxes. In contrast, one-stage regression-based methods, such as the center-based method of the YOLO series \cite{redmon2016you}, consider the center pixel of an object as a positive value and predict the distance from the positive value to the boundary of the bounding box. Recently, one-stage detectors have gained significant attention due to their surprising advantages over traditional two-stage detectors. Specifically, one-stage detectors \cite{jiang2022review,hurtik2022poly} require only one forward pass, making them faster and more suitable for real-time applications. Additionally, they do not need to generate candidate boxes, which reduces the amount of calculation and memory consumption. Therefore, this work builds on the recent real-time detector YOLOX \cite{ge2021yolox}, which is capable of balancing both speed and performance.

\subsection{Temporal detection methods}
Temporal detection methods exploit the inter-frame information to improve the performance and speed of the detectors in videos. Existing approaches can be divided into two types: feature-level and box-level. Feature-level methods leverage attention mechanisms \cite{guo2022uncertainty}, optical flow \cite{li2023sodformer}, and tracking methods \cite{cao2023towards}, aiming to aggregate rich features for complex video changes. Conversely, box-level methods \cite{pathiraja2023multiclass, shen2022confidence} aim to refine detection boxes by predicting temporal associations of bounding boxes during post-processing. This work proposes an online detector that endows spatial convolutions with temporal modeling capabilities to enrich temporal information comprehensively. Unlike existing approaches, our self-correcting detector optimizes feature extraction, fusion, and candidate box screening from both temporal and spatial levels.

\section{Methods}
\label{sec3}

\begin{figure}[!t]
	\centering
	\includegraphics[width=\textwidth]{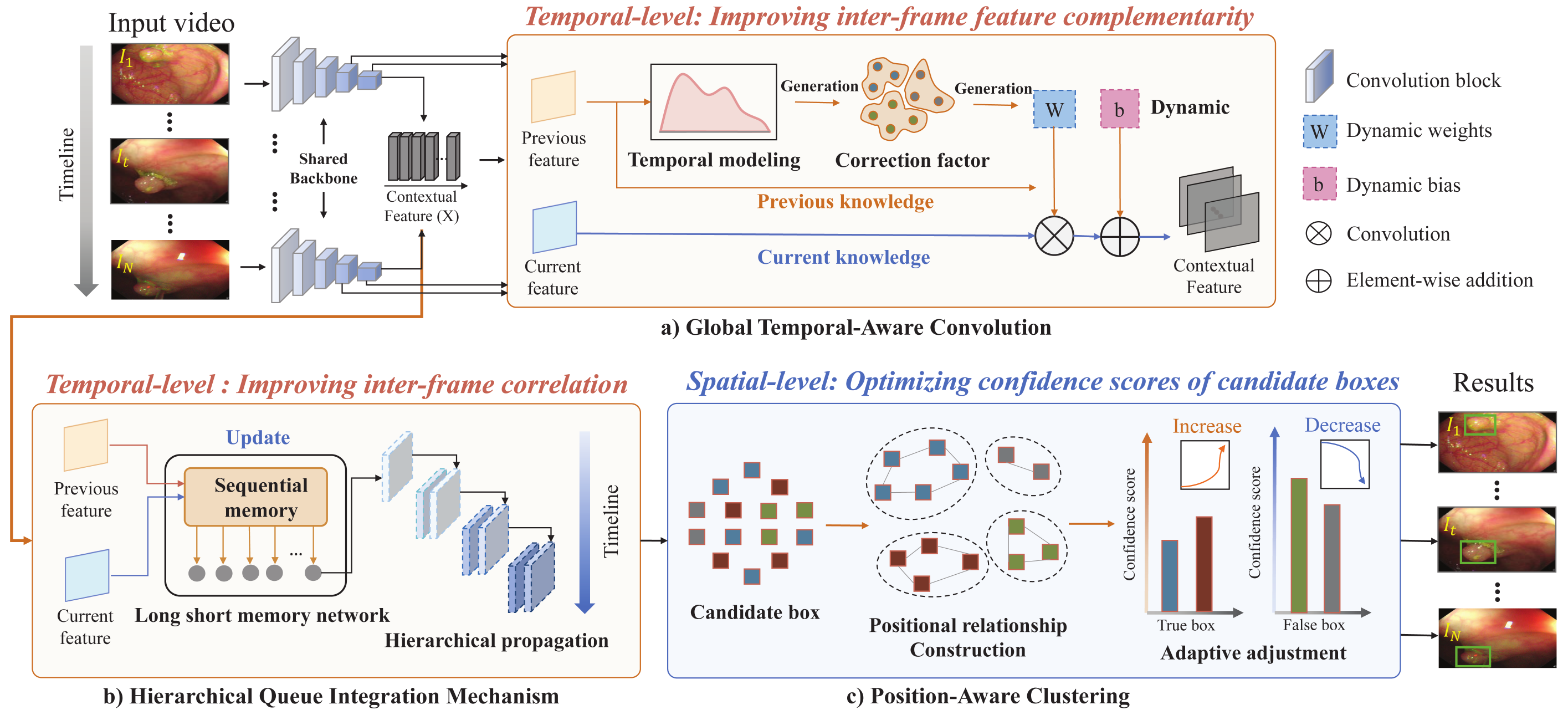}
	\caption{The overview of temporal-spatial detector architecture consists of temporal-level consistency learning and spatial-level reliability learning. a) \& b) At the temporal level, we aim to enhance the flexibility of feature extraction and fusion, thereby generating more reliable proposals. c) At the spatial level, we aim to reduce discrepancies between the confidence scores and the actual positive probabilities of candidate bounding boxes.}
	\label{img5}
\end{figure}

TSdetecter is a collaborative learning network that effectively leverages contextual information within spatial and temporal domains to enhance video polyp detection. In detail (Fig. \ref{img5}), the Global Temporal-aware Convolution (Section \ref{3.1}) calibrates the features obtained through convolution by generating dynamic weights guided by the previous features in the backbone stage. Subsequently, these calibrated features are propagated to the neck stage of the network. The Hierarchical Queue Integration Mechanism (Section \ref{3.2}) facilitates enhanced information integration across nearby frames, employing memory-based mechanisms and hierarchical propagation to fuse temporal information effectively. Finally, the Position-Aware Clustering (Section \ref{3.3}) further enhances predictions by meticulously considering the spatial relationships among the detected objects.

\textbf{Revisit the one-stage detector.} One-stage detectors encompass diverse module configurations while adhering to a fundamental architecture \cite{redmon2016you, redmon2018yolov3}. The architecture can be briefly outlined in four key components: the backbone, neck, detection head, and post-processing (Fig. \ref{img2}). The backbone network extracts feature maps from the input images. Subsequently, these feature maps are passed to the neck module, facilitating the aggregation of multi-level features. The detection head operates on all feature levels to generate predictions. Lastly, results are obtained through post-processing techniques, such as NMS. In this study, the YOLOX \cite{ge2021yolox} is chosen as the base network. It incorporates several innovative techniques, including decoupling headers and advanced label assignment, rendering it a formidable contender among real-time detectors. 

\subsection{Global Temporal-aware Convolution} \label{3.1}

Global temporal-aware convolution (GT-Conv) is designed to dynamically adjust convolution kernel weights based on temporal context, aiming to improve the model's capacity to capture and represent temporal patterns. To illustrate the differences between standard convolution, Traditional dynamic convolution, and GT-Conv, we present a visual comparison in Fig. \ref{img6}. The analysis highlights the following key observations: 1) Standard convolution uses fixed kernel weights \cite{chen2020dynamic}, limiting its adaptability to changes in input data. 2) Traditional dynamic convolution cannot to integrate temporal context knowledge, which is essential for obtaining inter-frame correlation in video detection tasks \cite{huang2021tada, hu2018squeeze}. In response to these limitations and drawing inspiration from the inherent calibration performed by temporal convolution, our GT-Conv consists of two basic steps: modeling of temporal information and generation of dynamic correction factors.

\subsubsection{The Dynamic Convolution Layer} 
Given the input feature $X_t$ of the current frame $I_t$ in the convolutional layer, traditional 2D convolution the output feature ${\bar{X}}_t$ can be obtained as follows:
\begin{equation}
{\bar{X}}_t = X_t * W_t + b_t 
\end{equation}
Among them, the operator $*$ represents the convolution operation. $W_t$ and $b_t$ are the weights and biases learned in the training, and they are shared throughout the feature extraction stage. Differently, ${\bar{W}}_t$ and ${\bar{b}}_t $ in the GT-Conv process are dynamically generated by the correction factor $\alpha_{t}^{w}$ and $\alpha_t^b$. It is worth noting that these correction factors vary from frame to frame, guided by previous frames' features, making the correction factors unique for each frame. The output features are as follows:
\begin{equation}
	{\bar{W}}_t = W_t * \alpha_{t}^{w}, {\bar{b}}_t = b_t * \alpha_t^b
\end{equation}
\begin{equation}
	{\bar{X}}_t = X_t * {\bar{W}}_t + {\bar{b}}_t 
\end{equation}

\subsubsection{The Modeling of Temporal Information}

The input is a contextual feature sequence $\bar{X}_t = \underbrace{\left\{X_{t-p},X_{t-q},…,X_t \right\}}_{k} $ of length $k$ including the current frame $X_t$ and the previous frames. Here, $p, q \leq t$ represent two separate indices of the frame before the current frame $X_{t}$, respectively. The reason why two separate symbols represent it is that the previous frames here are randomly selected from all frames before the current frame, and they may not necessarily be adjacent. To capture the inter-frame temporal dynamics while ensuring an adequate field of view, Global Average Pooling (GAP) and Spatial Attention Pooling (SAP) are employed on the feature maps  (Fig. \ref{img5}). Subsequently, each pooled feature element is aggregated separately by 3D convolution $F$ and addition with a kernel of 1 to generate a specific representation $\hat{S}_t$ as follows:
\begin{equation}
	\hat{S}_t=BN(ReLU(F(GAP(\hat{X}_t))+(F(SAP(\hat{X}_t))))
\end{equation}
In the equation, $ReLU$ denotes the ReLU activation function, while $BN$ represents batch normalization, both of which are employed to ensure the effectiveness and stability of the representation generation process.

\begin{figure}[!t]
	\centering
	\includegraphics[width=0.8\textwidth]{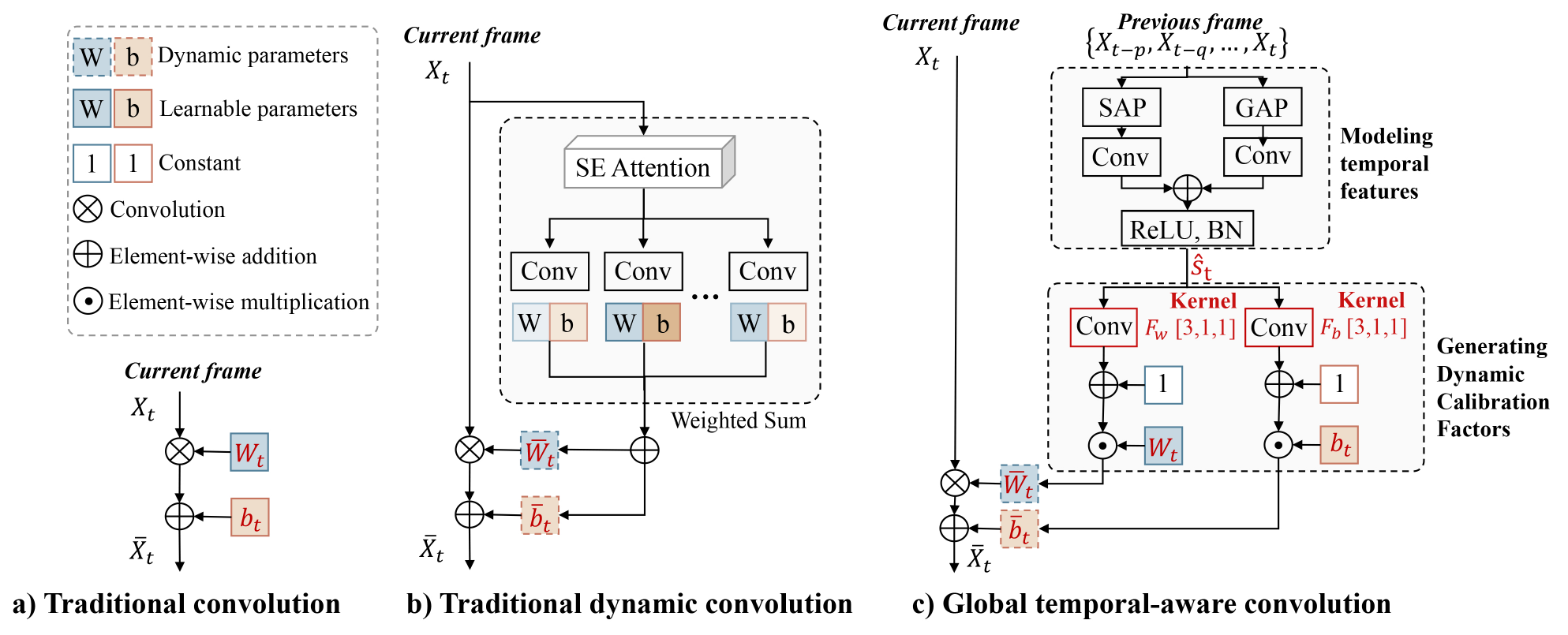}
	\caption{Global temporal-aware convolution differs from conventional convolutions in that its parameters can be adaptively adjusted in each frame. The temporal calibration factor is generated from the feature sequence of previous frames.}
	\label{img6}
\end{figure}

\subsubsection{The Generation of the Dynamic Correction Factor} Distinct correction factors are assigned to individual frames, thereby assigning unique weights and biases to each frame (Fig. \ref{img6}).  After obtaining the temporal feature expression $\hat{S}_t$, the fusion expression is achieved through a 3D convolution operation. The calibration factors $\alpha_{t}^{w}$ and $\alpha_t^b$ are generated as follows:
\begin{equation}
	\alpha_{t}^{w} = 1+F_{w} (\hat{S}_t), \alpha_t^b = 1+F_{b}(\hat{S}_t)
\end{equation}
where $F_{w}$ and $F_b$ represent three-dimensional convolutions with a kernel size of [3, 1, 1], where the temporal dimension is 3. The convolution operation is performed across the temporal dimension, capturing the temporal context provided by the previous frame sequence, essentially serving as a basic fundamental transformation within the model. Notably, at the initial stage of the model, $\alpha_{t}^{w}$ and $\alpha_t^b$ are set to 1; that is, the default weights and biases of the pre-training are loaded.

\textbf{Summary of the advantages.} The adaptability of kernel weights in the GT-Conv is achieved through the generation of dynamic correction factors, which are influenced by the temporal context information provided by the input frames. Benefiting from this capacity, the evolving inter-frame features are effectively exploited. To our knowledge, GT-Conv is the first successful attempt to integrate temporal information into feature extraction within a one-stage detector.

\subsection{Hierarchical Queue Integration Mechanism} \label{3.2}
\begin{figure}[!t]
	\centering
	\includegraphics[width=0.6\textwidth]{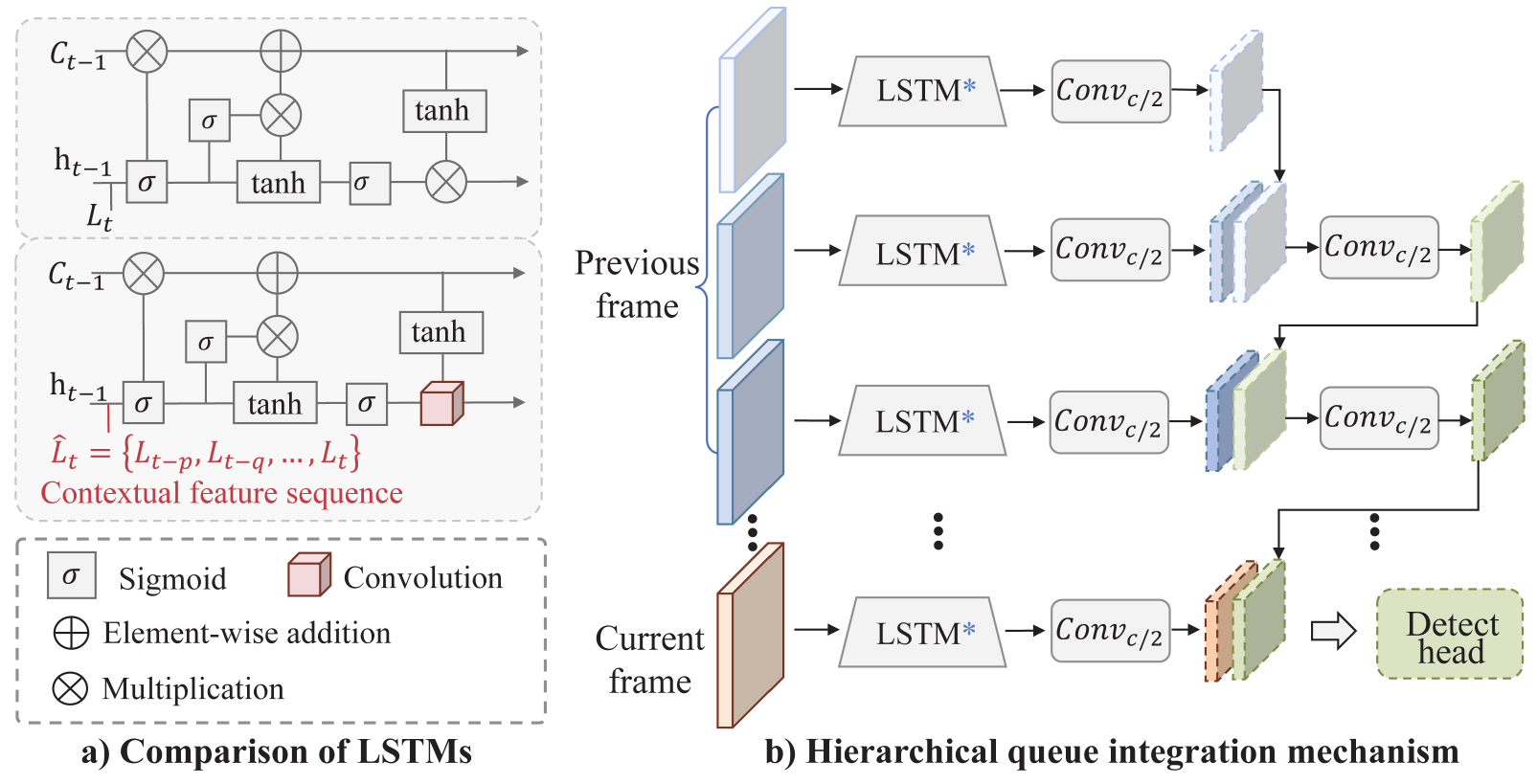}
	\caption{a) Unlike traditional LSTM, the proposed method can capture long memory and learn temporal correlations. b) The overall architecture of the memory interaction stream. After the neck stage, the memory interaction network is employed to capture long-term dependencies and temporal relationships among frames. By directly aggregating the features stored at the previous time step, the consistency between frames is maintained, leading to improved prediction robustness.}
	\label{img7}
\end{figure}

The hierarchical queue integration mechanism (HQIM) is designed to facilitate the continuous memory and integration of features across frames (Fig. \ref{img7}). This process ensures the seamless integration of features across frames, allowing for comprehensive processing of dynamic and temporal relationships in the data, thereby enhancing feature representation for more accurate object detection. The traditional neck stage often struggles to effectively utilize the previous information, and capture the dependency relationship between frames. Thus, HQIM harnesses and improves LSTM networks \cite{hochreiter1997long} to encode global contextual information efficiently and introduces a hierarchical propagation mechanism. The decision to use LSTM instead of a transformer structure is based on two factors: temporal modeling capabilities and computational efficiency. LSTMs excel at capturing temporal dependencies and generally require less data to train effectively compared to Transformers, making them suitable for video polyp detection.

The feature representation $L_t$ of the current frame is obtained from the neck stage and then passing through the traditional LSTM, can be expressed as follows:
\begin{equation}
	Input \, gate: f_t=\sigma(W_f L_t+W_f h_{t-1}+b_f)
\end{equation}
\begin{equation}
	Forget \, gate: i_t=\sigma(W_i L_t+W_i h_{t-1}+b_i)
\end{equation}
\begin{equation}
	Output \, gate: o_t=\sigma(W_o L_t+W_o h_{t-1}+b_o)
\end{equation}
\begin{equation}
	Input \, modulation \, gate: \tilde{C}_t=tanh(W_C L_t+W_C h_{t-1}+b_C)
\end{equation}
\begin{equation}
	C_t=f_t \cdot C_{t-1}+i_t \cdot \tilde{C}_t
\end{equation}
\begin{equation}
	h_t=o_t \cdot tanh \, C_t
\end{equation}
where $W$ and $b$ represent weight and bias, respectively. The function $\sigma$ denotes the sigmoid activation function, while $\cdot$ represents pointwise multiplication. $L_t$ corresponds to the input at time $t$. $h_t$ and $h_{t-1}$ is the output at time $t$ and $t-1$. 

 Unlike the traditional LSTM structure (Fig. \ref{img7} a), the original structure has been adapted to suit the detection task for polyp videos with three key modifications. First, the input $L_t$ is substituted with a contextual feature sequence $\hat{L}_t = \underbrace{\left\{L_{t-p},L_{t-q},…,L_t \right\}}_{k} $ of length $k$ to achieve a continuous feature representation. Second, The Hadamard product is replaced with a convolution operation, which allows for more effective extraction of spatial representations from the feature sequence. Third, the tanh function is replaced by a convolution operation when calculating the output $h$. The modified formula in this context is as follows:
\begin{equation}
	Input \, gate: f_t=\sigma(W_f * [\hat{L}_t, h_{t-1}] + b_f)
\end{equation}
\begin{equation}
	Forget \, gate: i_t=\sigma(W_i * [\hat{L}_t, h_{t-1}] + b_i)
\end{equation}
\begin{equation}
	Output \, gate: o_t=\sigma(W_o * [\hat{L}_t, h_{t-1}] + b_o)
\end{equation}
\begin{equation} 
	Input \, modulation \, gate: \tilde{C}_t=tanh(W_C * [\hat{L}_t, h_{t-1}] + b_C)
\end{equation}
\begin{equation}
	C_t=f_t \cdot C_{t-1}+i_t \cdot \tilde{C}_t
\end{equation}
\begin{equation}
	\hat{h}_t=F(o_t \cdot C_t)
\end{equation}
In practice, we employ convolution layer $F$ with kernel size 3 to aggregate the corresponding features.

After acquiring the context-enhanced feature $\hat{h}_t = \underbrace{\left\{L_{h-p},L_{h-q},…,h_t \right\}}_{k} $, the progressive accumulation mechanism is applied. This involves performing a $3 \times 3$ convolution on the motion features from the previous and current moments, reducing the number of channels by half. Note that motion features refer to spatiotemporal features extracted from nearby frames, as endoscopic videos involve continuous movement of the probe and exhibit continuous motion within the captured frames. The resulting modulated features are then concatenated. This step-by-step process is repeated until the features from all stages converge to $M_t$. The calculation formula for this process can be expressed as follows:
\begin{equation}
	M_t=Concat(F_{c/2} (Concat(F_{c/2} (h_{t-p} ),F_{c/2} (h_{t-p} ))),…,F_{c/2} (h_t ))
\end{equation}
where $F_{c/2}$ represents a convolutional layer that reduces the number of channels by half. Through the selective and comprehensive aggregation of multi-temporal features, an informative and fine-grained feature representation $M_t$ is encoded. This final feature representation serves as guidance for the detection head in the subsequent stages of the task.

\textbf{Summary of the advantages.} The temporal memory mechanism and progressive accumulation mechanism enable HQIM to integrate multi-moment consistant context. To  our knowledge, this is the first exploration of integrating temporal information into the detector neck stage.

\subsection{Position-aware Clustering} \label{3.3}
\begin{figure}[!t]
	\centering
	\includegraphics[width=\textwidth]{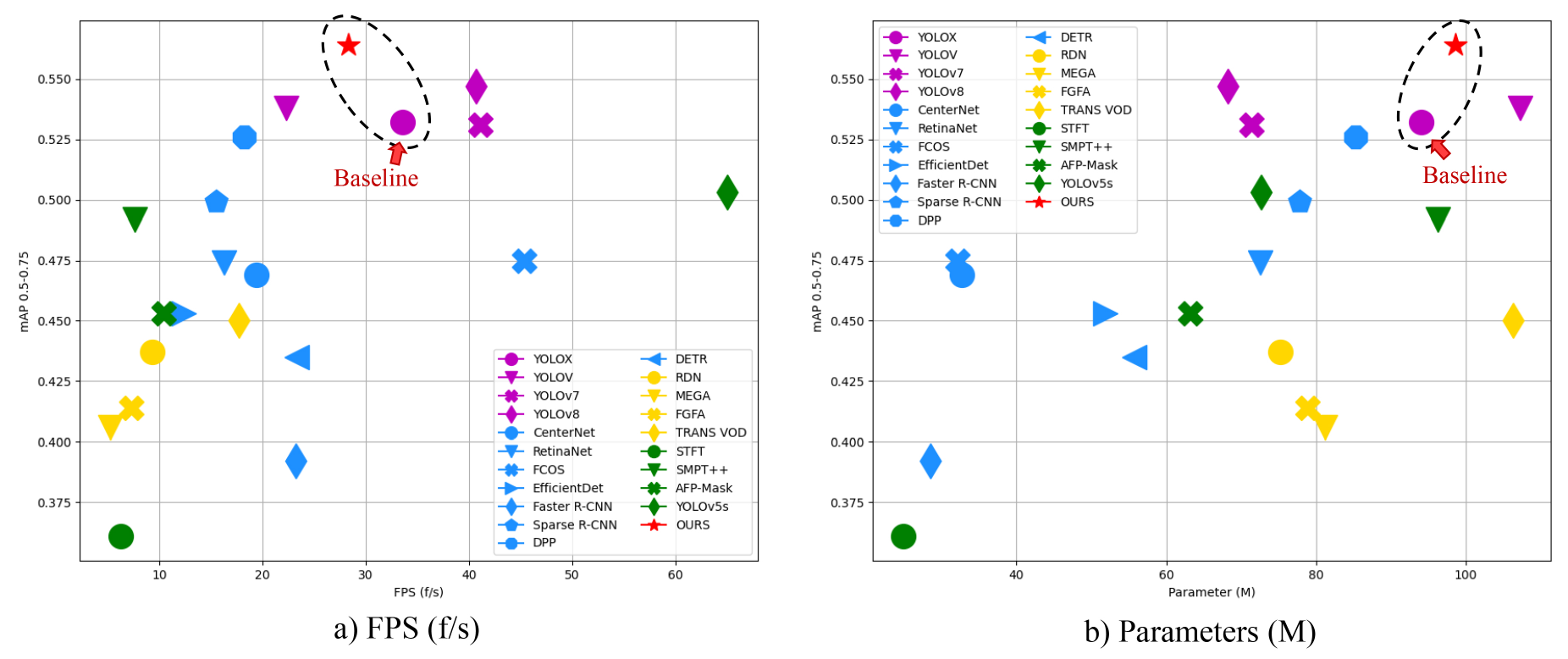}
	\caption{Position-aware clustering refines the confidence scores during the post-processing phase by considering the confidence of adjacent bounding boxes. All candidate boxes within the graph are initially transformed into a spatial relationship graph during this process. Positive samples strengthen the confidence, while negative samples diminish it, thereby facilitating reliable predictions.}
	\label{img8}
\end{figure}

The position-aware clustering (PAC) reduces the difference between the confidence score of the candidate bounding box and the true positive probability by considering the proximity relationship of the bounding box (Fig. \ref{img5}). Simply put, it is based on the idea that if the adjacent bounding box has a high confidence score, the candidate box is also more likely to be valid, and vice versa. Specifically, PAC (Fig. \ref{img8}) consists of three key steps: construction of position relationships, enhancement of positive sample candidates, and suppression of negative sample candidates. By integrating these three steps, the spatial dependence between adjacent boxes can be effectively exploited to adjust the confidence of candidate boxes dynamically.

\subsubsection{Construction of Positional Relationship} This step establishes the positional relationship between adjacent candidate bounding boxes based on the Intersection over Union (IoU) metric. The process is illustrated in Fig. \ref{img8}. Given an image, the bounding boxes before post-processing are combined and denoted as $\gamma=\left\{a_1,a_2,…,a_n\right\}$. For each pair of bounding boxes $a_i,a_j \in \gamma$ with an IoU greater than a threshold value $\theta$, they are considered adjacent pairs. Subsequently, a relationship graph $\Omega=\left\{b_1,b_2,…,b_n\right\}$ is constructed for each image. For a specific box $b_i \in \Omega$, the edge set $\varepsilon_{b_i} $ represents the edges connecting the bounding boxes, and the node set $\nu_{b_i} $ represents the individual bounding boxes. For a box $a_i \in \nu_{b_i} $, the neighbor node set $N_{a_i}$  comprises all the nodes connected to $a_i$ in the graph $\Omega$.

\subsubsection{Enhancement of Positive Sample Candidates} In this step, the confidence scores of the positively classified candidate bounding boxes are increased, which accurately represent samples of the object of interest. Specifically, for a given bounding box $a_i$, positive information is generated from its neighbor nodes $N_{a_i}$ to enhance its confidence score $P(a_i)$. It is assumed that neighbors with lower confidence scores can provide evidence of true confidence. For a bounding box $a_j \in \nu_{b_i} $, its set of low neighbors $L_{a_i} $ is a subset of its neighbors $N_{a_i}$ . The inclusion criteria for $L_{a_i} $ are as follows: the $IoU (a_i, a_j)$ is greater than a threshold value $\delta$, and the confidence score $P(a_j)$ is lower than $P(a_i)$. Typically, the threshold value $\delta$ is set to be larger than the neighbor threshold $\theta$. Finally, the positive enhancement value $E(a_j)$ of the confidence score for the candidate bounding box $a_i$ can be computed as follows:
\begin{equation}
	E(a_j)=\frac{Q}{Q+1} \cdot (1-P(a_i )) \cdot \underset{a_j \in L_{a_i}}{max} P(a_j)
\end{equation}
where $Q$ is the number of low neighbors $L_{a_i}$ in this set.

\subsubsection{Suppression of Negative Sample Candidates} Conversely, for a bounding box  $a_j \in \nu_{b_i} $, its high neighbor $H_{a_i}$ satisfies the condition $IoU (a_i, a_j) > \delta $ and $P(a_j) > P(a_i)$. If a high-confidence neighbor $H_{a_i} $ exists, the confidence score of the current box $a_i$ will be suppressed. In this case, the bounding box $a_j$ with the highest confidence value in $H_{a_i}$ is selected to suppress $a_i$. Therefore, the negative suppression value $S(a_j)$ for the candidate bounding box $a_i$ can be calculated as follows:
\begin{equation}
	S(a_j)= P(a_i ) \cdot IoU (a_i,a_j)
\end{equation}
where $a_j$ is the highest confidence value among neighbors. Finally, the confidence $\hat{P}(a_i )$ after the correction of $a_i$ is:
\begin{equation}
	\hat{P}(a_i ) = P(a_j )+E(a_j)-S(a_j)
\end{equation}

\textbf{Summary of the advantages.} Relying on position-aware clustering to improve the NMS method in the post-processing stage leverages the spatial relationships among candidate bounding boxes and efficiently eliminates redundant bounding boxes. To the best of our knowledge, PAC is the first attempt to adaptively optimize confidence from a clustering approach based on belief propagation.

\section{Experiments configuration }
\label{sec4}
\subsection{Datasets}
The proposed method is comprehensively evaluated on SUN, CVC-ClinicDB, and PICCOLO datasets. 1) The SUN dataset \cite{misawa2021development} \href{http://amed8k.sundatabase.org/}{Link} is the largest benchmark for video polyp detection and encompasses a total of 112 cases, consisting of 100 positive cases (with polyps) and 12 negative cases (without polyps). The positive cases contain 49138 frames and are partitioned into training (32343 images), validation (5181 images), and test sets (11611 images) using a ratio of 7:1:2, respectively. Notably, the division is performed per-case basis, ensuring that each case appears in only one of the sets. The negative cases are tested to evaluate the model's ability to combat false positives. 2) The CVC-ClinicDB dataset \cite{bernal2015wm} \href{https://polyp.grand-challenge.org/CVCClinicDB/}{Link} focuses on image-based polyp detection, comprising 612 images.  Following official guidelines, the dataset is partitioned into a training set of 550 images and a test set of 62 images. 3) The PICCOLO dataset \cite{sanchez2020piccolo} \href{https://www.biobancovasco.org/en/Sample-and-data-catalog/Databases/PD178-PICCOLO-EN.html}{Link} encompasses 3433 images extracted from clinical colonoscopy videos involving 48 patients. Officially divided, the dataset comprises 2203 images for training, 897 for validation, and 333 for testing. 
  
\subsection{Evaluation metrics}
The evaluation of the model encompasses three main aspects: detection box performance, classification performance, and speed performance. Regarding detection box accuracy, COCO evaluation \cite{lin2014microsoft} is adopted, which is a standard benchmark in the field of object detection. The average precision (AP) is calculated for IoU thresholds ranging from 0.5 to 0.95. Additionally, $AP_m$ and $AP_l$ are reported, representing the average precision for medium and large objects, respectively. In terms of classification performance, the model is assessed using metrics such as mean Average Precision (mAP), Precision, Recall, and F1-score. It should be noted that the concept of recall used in this study is the same as the Sensitivity metric, and it is uniformly expressed as Recall here. Lastly, the model's ability to balance computational performance, number, and speed when processing video is evaluated by measuring IoU, number of parameters, and frames per second (FPS), respectively.

\subsection{Implementation details} All the experiments are fine-tuned from the COCO pre-trained model by 50 epochs. The training is conducted on an NVIDIA GeForce RTX 3090 GPU, with a batch size set to 2. For training, we employ stochastic gradient descent (SGD) as the optimization algorithm and adopt a learning rate of $0.001\times batch size / 64$ (linear scaling) and the cosine schedule with a warmup strategy for 5 epoch. The weight decay is set to 0.0005, and the SGD momentum is set to 0.937. The input size of the image during training is set to $640 \times 640$. Data augmentation techniques are consistent with the base model, including: Mosaic and Mixup \cite{ge2021yolox}. The length of the contextual feature sequence, including the current frame and the previous frames, is set to 4. In addition, all training settings of other methods are implemented according to the optimal configurations mentioned in their respective papers.

\section{Results and Discussion}
\label{sec5}
\subsection{Comparative with Existing Methods}

\begin{table}[!] 
	\scriptsize
	\setlength{\belowcaptionskip}{5pt}
	\caption{Overall performance of all four types of 20 detection frameworks tested on the \textbf{SUN} colonoscopy video dataset. "Type" represents the category of the method being compared.} The best results are highlighted in bold.
	\centering
	\renewcommand\arraystretch{1.4}
	\resizebox{0.9\textwidth}{!}{
		\begin{tabular}{cc|ccccccccc}
			\hline
			Method&Type&$AP_{0.5-0.75}$&	$AP_{0.5}$&	$AP_{0.75}$&$	AP_{m}$&$	AP_{l}$&	mAP&	Precision&	Recall&	F1\\
			\hline
			YOLOX&\multirow{4}*{YOLO-Based}&	0.524&	0.937&	0.545&	0.339&	0.512&	0.937&	0.895&	0.904&	0.910 \\
			YOLOV&&	0.538&	0.953&	0.563&	0.368&	0.542&	0.945&	0.882&	0.904&	0.892\\
			YOLOv7&&	0.531&	0.934&	0.545&	0.195&	0.537&	0.940&	0.941&	0.812&	0.878 \\
			YOLOv8&	&0.547&	0.945&	0.582&	0.271&	0.552&	0.949&	0.954&0.836&	0.892\\
			\hline
			CenterNet&\multirow{8}*{Image-Based}&	0.469&	0.886&	0.435&	0.192&	0.473&	0.866&\textbf{	0.992}&	0.410&	0.590 \\
			RetinaNet&&	0.474&	0.894&	0.466&	0.123&	0.477&	0.898&	0.837&	0.855&	0.853 \\
			FCOS&&	0.475&	0.926&	0.423&	0.166&	0.479&	0.931&	0.915	&0.878&0.901 \\
			EfficientDet&&0.499&	0.904&	0.513&	0.258&	0.502&	0.909&	0.926&	0.854	&0.896 \\
			Faster R-CNN&&	0.472&	0.893&	0.386&	0.122&	0.475&	0.907&	0.543&	0.847&0.665  \\
			Sparse R-CNN&&0.499&	0.904	&0.513	&0.258&	0.502&	0.889&	0.926&	0.854&	0.895\\									
			DPP	&&0.526&	0.911&	0.560&	0.163&	0.529&	0.920&	0.968&	0.749&	0.851 \\
			DETR&&	0.435&	0.892&	0.371&	0.052&	0.440&	0.897&	0.779&	0.898&	0.836\\
			\hline
			RDN&\multirow{4}*{Video-Based}&	0.437&	0.894&	0.371&	0.051&	0.440&0.899&	0.720&	0.899&	0.858\\
			MEGA&&	0.406&	0.855&	0.348&	0.060&	0.410&	0.860&	0.603&	0.885&	0.728\\
			FGFA&&	0.414&	0.843&	0.359&	0.018&	0.422&	0.848&	0.720&	0.855&	0.784\\
			TRANS VOD&&	0.450&	0.904&	0.382&	0.310&	0.452&	0.910&	0.932&	0.914&\textbf{	0.927}\\
			
			\hline										
			STFT&\multirow{4}*{Polyp-Based}&0.361&	0.807&	0.255&	0.029&	0.364&	0.712&	0.799&	0.794&	0.805 	\\	
			SMPT++&&0.492&	0.891&	0.471&	0.212&	0.489&	0.904&	0.921&	0.832&	0.866\\
			AFP-Mask&&0.453&	0.885&	0.409&	0.072&	0.457&	0.890&	0.911&	0.762&	0.830\\
			YOLOv5s &&	0.503&	0.893&	0.516&	0.186&	0.509&	0.907&	0.923&	0.839&	0.874\\
			\hline										
			Ours&Hybrid&	\textbf{0.564}&	\textbf{0.953}&\textbf{	0.612}&	\textbf{0.384}&\textbf{
			0.566}&	\textbf{0.954}&	0.910&	\textbf{0.928}&	0.921\\
			\hline
	\end{tabular}}
	\label{tab2}
\end{table}

\begin{table}[!] 
	\scriptsize
	\setlength{\belowcaptionskip}{5pt}
	\caption{Overall performance of all four types of 20 detection frameworks tested on the \textbf{CVC-ClinicDB} colonoscopy dataset. "Type" represents the category of the method being compared. The best results are highlighted in bold.}
	\centering
	\renewcommand\arraystretch{1.4}
	\resizebox{0.9\textwidth}{!}{
		\begin{tabular}{cc|cccccccccc}
			\hline
			Method&Type&$AP_{0.5-0.75}$&	$AP_{0.5}$&	$AP_{0.75}$&$	AP_{s}$&$	AP_{m}$&$	AP_{l}$&	mAP&	Precision&	Recall&	F1\\
			\hline
			YOLOX&\multirow{4}*{YOLO-Based}&0.712&	0.882&	0.794&	0.633&	0.694&	0.774&	0.893&	0.910&	0.889&	0.941\\
			YOLOV&&0.723&	0.884&	0.800&	0.630&	0.676&	0.794&	0.881&	0.904&	0.875&	0.940\\
			YOLOv7&&0.738&\textbf{	0.907}&	0.812&	0.573&	0.702&0.828&	0.903&	0.918&0.875&\textbf{	0.956}\\
			YOLOv8&	&0.739&	0.883&	0.808&	0.578&0.690&\textbf{	0.843}&	0.884&	0.910&	0.884&	0.940\\
			\hline
			CenterNet&\multirow{8}*{Image-Based}&0.682&	0.870&	0.771&	0.454&	0.661&	0.755&0.767&	0.912&	0.875&0.955 \\
			RetinaNet&&	0.630&	0.871&	0.721&	0.375&0.603&	0.707&	0.866&	0.895&	0.861&	0.912\\
			FCOS&&0.715&	0.879&	0.796&	0.526&0.670&	0.803&	0.884&	0.924&	\textbf{0.903}&	0.928 \\
			EfficientDet&&0.682&	0.872&	0.769&0.597&	0.641&	0.762&	0.874&	0.885&0.875&	0.940 \\
			Faster R-CNN&&	0.672&0.887&	0.795&	0.234&	0.633&	0.777&	0.888&	0.896&	0.889&	0.889 \\
			Sparse R-CNN&&0.686&	0.890&	0.780&	0.393&	0.662&	0.754&	0.896&	0.903&	0.875&	0.926\\									
			DPP	&&0.687&0.903&	0.796&	0.423&	0.631&0.753&	0.899&	0.905&	0.875&	0.926 \\
			DETR&&0.472&	0.828&	0.575&	0.136&0.402&	0.638&	0.838&	0.847&	0.847&	0.836\\
			\hline
			RDN&\multirow{4}*{Video-Based}&	0.682&0.849&	0.762&	0.507&	0.626&	0.759&0.862&	0.877&	0.863&	0.895\\
			MEGA&&0.656&	0.825&	0.736&	0.489&	0.623&0.701&	0.831&	0.849&	0.832&	0.867\\
			FGFA&&	0.647&	0.813&0.729&	0.456&0.620&	0.702&	0.826&	0.842&	0.829&	0.862\\
			TRANS VOD&&0.685&	0.847&0.763&	0.521&0.679&	0.762&	0.866&	0.878&	0.866&	0.898\\
			
			\hline										
			STFT&\multirow{4}*{Polyp-Based}&0.645&0.825&	0.717&	0.486&	0.647&	0.711&	0.848&0.825&	0.840&	0.826	\\	
			SMPT++&&0.654&	0.830&	0.723&0.493&0.655&0.721&	0.851&	0.870&	0.845&	0.902	\\
			AFP-Mask&&0.658&	0.832&0.726&	0.495&	0.657&	0.723&	0.854&	0.872&	0.848&	0.904\\
			YOLOv5s &&0.699&	0.867&	0.772&0.521&	0.688&	0.764&	0.885&	0.902&	0.875&	0.928\\
			\hline										
			Ours&Hybrid&\textbf{0.745}&	0.898&\textbf{	0.832}&\textbf{	0.657}&\textbf{	0.723}&	0.796&\textbf{	0.905}&	\textbf{0.926}&	0.893&0.949\\
			\hline
	\end{tabular}}
	\label{clinicdb}
\end{table}

\begin{table}[!] 
	\scriptsize
	\setlength{\belowcaptionskip}{5pt}
	\caption{Overall performance of all four types of 20 detection frameworks tested on the \textbf{PICCOLO} colonoscopy dataset. "Type" represents the category of the method being compared. The best results are highlighted in bold.}
	\centering
	\renewcommand\arraystretch{1.4}
	\resizebox{0.9\textwidth}{!}{
		\begin{tabular}{cc|ccccccccc}
			\hline
			Method&Type&$AP_{0.5-0.75}$&$AP_{0.5}$&$AP_{0.75}$&$AP_{m}$&$AP_{l}$&mAP&Precision&Recall&F1\\
			\hline
			YOLOX&\multirow{4}*{YOLO-Based}&0.625&0.858&0.681&0.278&0.668&0.773&0.918&0.615&	0.742\\
			YOLOV&&0.618&0.872&	0.684&0.278&0.658&0.783&0.883&0.685&0.773\\
			YOLOv7&&0.633&0.871&0.678&0.280&0.675&0.787&0.903&0.651&0.762 \\
			YOLOv8&&0.618&0.815&0.666&0.489&0.671&	0.739&0.861&0.612&	0.727\\
			\hline
			CenterNet&\multirow{8}*{Image-Based}&0.568&	0.802&	0.636&	0.456&	0.612&	0.728&	\textbf{0.963}&	0.564&0.713\\
			RetinaNet&&0.534&0.812&	0.604&	0.434&0.572&0.736&	0.851&0.651&0.744\\
			FCOS&&0.637&\textbf{0.907}&	0.682&0.522&	0.668&	0.796&	0.890&	0.675&	0.772 \\
		EfficientDet&&0.529&0.778&0.577&0.439&0.559&	0.707&0.909&	0.568&0.707\\
		Faster R-CNN&&	0.537&0.811&	0.607&	0.476&0.559&	0.737&	0.700&	0.728&	0.717 \\
		Sparse R-CNN&&0.616&0.877&	0.707&	0.541&	0.650&	0.800&	0.839&	0.757& 0.801\\									
			DPP	&&0.625&0.858&	0.681&	0.278&	0.668&	0.773&	0.918&	0.615&	0.750\\
		DETR&&0.389&0.764&0.356&	0.259&	0.436&	0.697&	0.701&0.704&	0.703\\
			\hline
		RDN&\multirow{4}*{Video-Based}&	0.558&	0.835&0.614&	0.439&	0.599&	0.757&	0.844&	0.665&	0.748\\
		MEGA&&0.600&	0.802&	0.629&	0.466&	0.640&0.731&0.930&0.586&0.722\\
		FGFA&&	0.622&	0.825&0.666&0.518&0.655&0.749&0.892&0.616&	0.736\\
		TRANS VOD&&	0.626&	0.853&	0.706&	0.511&	0.661&	0.776&	0.907&0.686&	0.781\\
			\hline										
		STFT&\multirow{4}*{Polyp-Based}&0.483&	0.706&	0.521&	0.416&	0.521&	0.641&	0.850&	0.546&	0.663	\\	
			SMPT++&&0.529&	0.771&	0.580&	0.434&0.576&	0.703&	0.932&	0.452&	0.615\\
		AFP-Mask&&0.519&	0.797&	0.575&	0.444&0.543&0.723&	0.640&	0.726&	0.690\\
		YOLOv5s &&0.571&	0.776&	0.622&	0.486&	0.616&	0.705&	0.862&	0.588&	0.706\\
			\hline										
			Ours&Hybrid&\textbf{0.657}&	0.885&\textbf{	0.732}&	\textbf{0.546}&\textbf{	0.707}&\textbf{	0.824}&	0.896&\textbf{0.776}&	\textbf{0.837}\\
			\hline
	\end{tabular}}
	\label{PICCOLO}
\end{table}

To provide a comprehensive evaluation, comparative experiments include 20 state-of-the-art detection methods in four types. 1) YOLO Series Methods (YOLOX \cite{ge2021yolox}, YOLOV \cite{shi2023yolov}, YOLOv7 \cite{wang2023yolov7}, YOLOv8 \href{https://github.com/ultralytics/ultralytics}{Link} ). Given that the backbone of the proposed framework is derived from YOLOX with further innovative enhancements, it is compared with more advanced iterations in the YOLO family. 2) Image-Based Detection Methods. These methods are widely used in medical image detection tasks and are often used as the backbone that can be modified for specific tasks. Our evaluation encompasses three architecture categories: single-stage methods (CenterNe \cite{duan2019centernet}, FCOS \cite{tian2019fcos}, RetinaNet \cite{lin2017focal}, EfficientDet \cite{tan2020efficientdet}), two-stage methods (Faster R-CNN \cite{girshick2015fast}, Sparse R-CNN \cite{sun2021sparse}, DPP \cite{li2022should}), and transformer-based methods (DETR) \cite{carion2020end}. 3) Video-Based Detection Methods (RDN \cite{deng2019relation}, MEGA \cite{chen2020memory}, FGFA \cite{zhu2017flow}, TRANS VOD \cite{zhou2022transvod}). This evaluation encompasses both popular residual and distillation networks, in addition to recent transformer-based architectures. 4) Polyp-Based Methods (STFT \cite{wu2021multi}, SMPT++ \cite{liu2022source}, AFP-Mask \cite{wang2022afp}, YOLOv5s \cite{karaman2023robust} ).  All these methods are experimented with using the same training configuration for fair comparison. 

\textbf{Comparative results on the SUN dataset.} TSdetector achieves the highest scores in various metrics (Table \ref{tab2}). Specifically, it attains the highest AP values at different IoU thresholds, including $AP_{0.5-0.75}$ (0.564), $AP_{0.5}$ (0.953), and $AP_{0.75}$ (0.612). These results confirm our expectation of exceptional detection rate and accuracy. In comparison to the cutting-edge YOLOv8 model, our method exhibits noteworthy improvements of 1.7\%, 0.8\%, and 3.0\% for these respective indicators. Moreover, in contrast to image-based methodologies, our proposed approach achieves a substantial enhancement in Recall, surpassing similar methods by a minimum margin of 3.8\% ($AP_{0.5-0.75}$). When contrasted with existing polyp detection techniques, our method surpasses them with $AP_{0.5-0.75}$  indicators showing 6.10\% to 20.3\% higher performance, demonstrating the superiority in colonoscopy polyp video detection scenarios. Compared with YOLOv5s, which achieved suboptimal results, TSdetector exhibited 6\% higher $AP_{0.5}$ and 9.6\% higher $AP_{0.75}$, emphasizing the potential to improve the accuracy of lesion identification.

\begin{table}[!] 
	\scriptsize
	\setlength{\belowcaptionskip}{5pt}
	\caption{Comparative quantitative results on IoU, parameters, and FPS for all four types of 20 detection methods on the SUN colonoscopy video dataset. The best results are highlighted in bold.}
	\centering
	\renewcommand\arraystretch{2.4}
	\resizebox{\textwidth}{!}{
		\begin{tabular}{c|cccccccccccccccccccc|c}
			\hline
			Method&YOLOX&YOLOV&YOLOv7&YOLOv8&CenterNet&RetinaNet&FCOS&EfficientDet&Faster R-CNN&SMPT+++&AFP-Mask&YOLOv5s&
			Sparse R-CNN&DPP&DETR&RDN&MAGA&FGFA&TRANS VOD&STFT&Ours\\
			\hline
			IoU&73.77&75.61&74.38&77.54&68.19&70.48&72.92&73.24&69.56&73.65&75.49&64.43&65.71&62.52&66.38&67.14&59.86&61.32&62.35&66.98&\textbf{80.73}\\									
			Parameters&92.07&107.26&71.34&68.23&32.66&72.60&32.15&51.84&28.48&77.85&85.32&55.68&75.24&81.26&78.93&\textbf{106.30}&24.89&96.33&63.24&72.68&98.61\\									
			FPS&33.61&22.34&41.10& \textbf{40.70}&19.43& 16.27& 45.33& 12.39&23.22&15.55&18.23&23.34&9.32& 5.28& 7.27& 17.76&6.26&7.67&10.43&65.05&28.29\\
			\hline
	\end{tabular}}
	\label{tab_iou}
\end{table}

\begin{figure}[!t]
	\centering
	\includegraphics[width=0.8\textwidth]{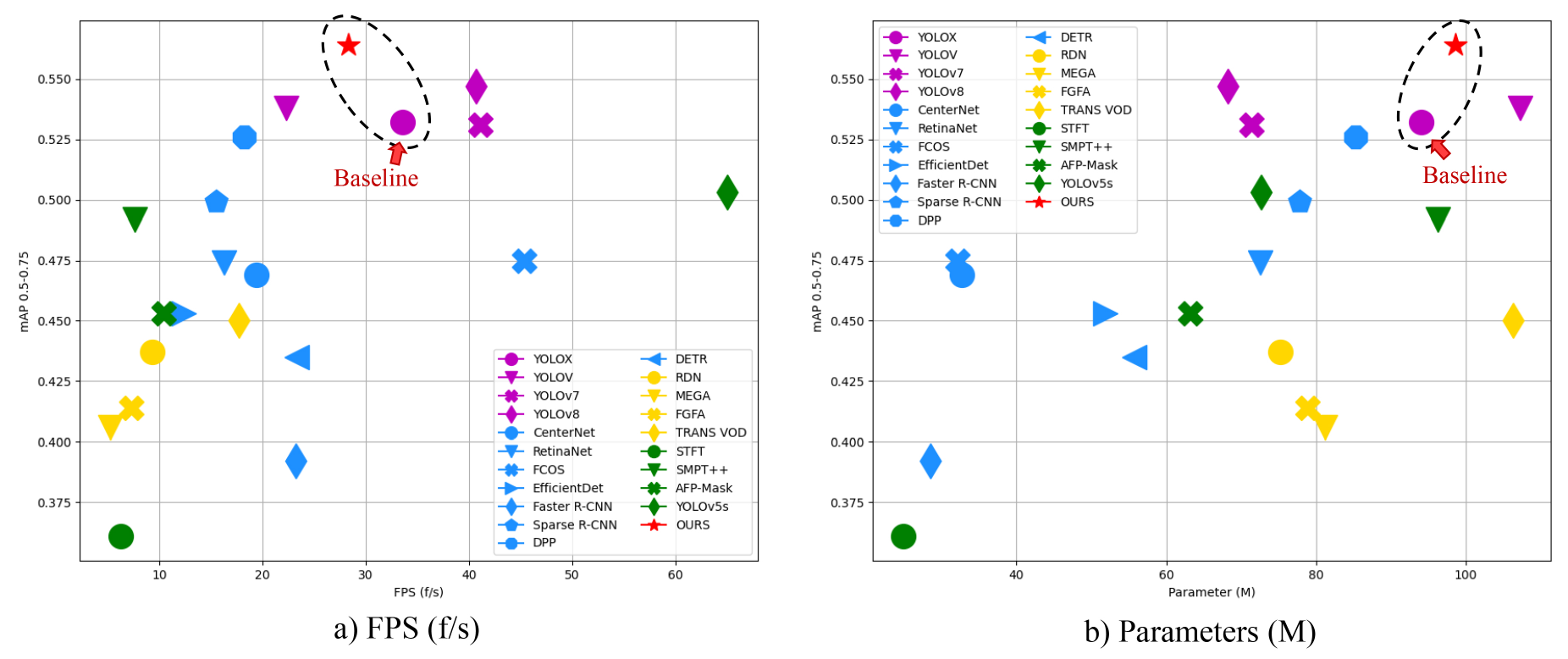}
	\caption{Compared to other object detectors in terms of FPS and number of parameters on the SUN dataset, TSdetector achieves the best trade-off between speed and accuracy.}
	\label{img9}
\end{figure}

\textbf{Comparative results on the CVC-ClinicDB dataset.} With a mAP of 90.5\% (Table \ref{clinicdb}), our method outperforms all others, showcasing its robustness in accurately detecting polyps across varying conditions. Moreover, achieving precision at 92.6\% and recall at 89.3\%, the TSdetector maintains high accuracy in identifying polyps while also capturing a high proportion of true positives. Comparative analyses against the YOLO-based, and other image-based, video-based, and polyp-based methods consistently highlight the superiority of our approach. These results collectively emphasize the potential of our hybrid method as an effective tool for computer-aided diagnosis.

\textbf{Comparative results on the PICCOLO dataset.} The comprehensive numerical analysis reveals significant performance disparities among the evaluated methods and the great potential of the proposed method in polyp detection (Table \ref{PICCOLO}). YOLOv7 leads the YOLO series with the highest $AP_{0.5-0.75}$ of 63.3\%, while FCOS exhibits strong performance within the Image-Based category, surpassing an $AP_{0.5-0.75}$ of 63.7\%. In the Video-Based category, TRANS VOD stands out with $AP_{0.5-0.75}$ scores with 62.6\%. Notably, our proposed hybrid framework achieves the highest $AP_{0.5-0.75}$ of 65.7\%, showcasing a substantial improvement over existing methods. Additionally, the approach strikes a balance between precision and recall, resulting in a commendable F1-score of 83.7\%, demonstrating robust performance across various evaluation metrics.
 
\textbf{Analysis of parameter number, speed, and performance.} The results reveal that our method achieves the highest IoU score of 80.73, indicating superior performance in object detection compared to the other models (Fig. \ref{img9}, Table \ref{tab_io}). Specifically, YOLOv8 achieved a sub-optimal IoU result of 77.54, but it is still 3.19 lower than the TSdetector. While current video-based methods yield commendable results, they often introduce computational overhead due to intricate temporal modules. In contrast, our method achieves a processing rate of 28.29 FPS, a significant advancement over existing video-based methodologies. In addition, the total number of parameters of the proposed method TSdetector is 98.61M, which is only a marginal increase of 6.54 parameters compared to the baseline YOLOX. These outcomes collectively exhibit the multifaceted advantages of our approach, trade-off both heightened accuracy and processing speed, thereby facilitating more precise lesion identification within intestinal endoscopic images.

\subsection{Ablation Study}

\begin{table}[!t] 
	\scriptsize  %%设置字体大小
	\setlength{\belowcaptionskip}{5pt}
	\caption{The ablation studies validate the effectiveness of the proposed modules on the SUN colonoscopy video dataset. GT-Conv, HQIM, and PAC represent the Global Temporal-aware Convolution,  Hierarchical Queue Integration Mechanism, and Position-Aware Clustering, respectively. The best results are highlighted in bold.}
	\centering
	\renewcommand\arraystretch{1.4}
	\resizebox{0.6\textwidth}{!}{
		\begin{tabular}{ccc|ccccc}
			\hline
			GT-Conv&HQIM&PAC&$AP_{0.5-0.75}$&	$AP_{0.5}$&	$AP_{0.75}$&$	AP_{m}$&$	AP_{l}$\\
			\hline
			&&&	0.524&	0.937&	0.545&	0.339&	0.512\\
			\checkmark&&&0.545&	0.937&	0.582&	0.364&	0.534\\
			&\checkmark&&0.544&	0.946&	0.575&	0.352&	0.548\\
			&&\checkmark&0.542&	0.941&	0.572&	0.355&	0.545\\
			\checkmark&\checkmark&&0.554&0.937&0.605&0.368&	0.558\\
			\checkmark&&\checkmark&0.552&0.934&0.608&0.356&0.556\\									
			\checkmark&\checkmark&\checkmark&\textbf{0.564}&\textbf{0.953}&	\textbf{0.612}&\textbf{0.384}&\textbf{0.566}\\
			\hline
	\end{tabular}}
	\label{tab3}
\end{table}

\textbf{Ablation for each submodule.} The original YOLOX-X model is the baseline, with each sub-module progressively incorporated: GT-Conv, HQIM, and PAC. The results highlight the crucial contributions of all sub-modules in achieving precise detection, as detailed in Table \ref{tab3}. Initially, the baseline model achieves 52.4\% accuracy at $AP_{0.5-0.75}$. Introducing GT-Conv leads to a noticeable performance improvement of 2.1\% compared to the baseline, underscoring the significance of weight calibration and the incorporation of temporal context. The integration of a more robust non-linear weight generation mechanism produces even more substantial performance enhancements. Subsequently, the addition of the HQIM underscores the substantial benefits of multi-frame information feature fusion, surpassing the performance achieved with single frames and leading to a 2.0\% increase in $AP_{0.5-0.75}$. Finally, lines 4, 6, 7 demonstrate the robust localization precision enhancement achieved by the PAC module. Comparing these results with those of lines 1, 2, 5 reveals improvements of 1.8\%, 0.7\%, and 1.0\%, respectively. Notably, the enhancements about $AP_{0.75}$ are even more pronounced, amounting to 2.7\%, 2.6\%, and 0.7\%, respectively. These findings further demonstrate the role of adaptive confidence in effectively guiding the precision of box-level detection.

\begin{table}[!t] 
	\scriptsize  %%设置字体大小
	\setlength{\belowcaptionskip}{5pt}
	\caption{The quantitative results of placing components in TSdetector  in different backbones on the SUN dataset illustrate the effectiveness of the proposed concept.}
	\centering
	\renewcommand\arraystretch{1.4}
	\resizebox{0.7\textwidth}{!}{
		\begin{tabular}{c|c|ccccc}
			\hline
			Method&Type&$AP_{0.5-0.75}$&	$AP_{0.5}$&	$AP_{0.75}$&$	AP_{m}$&$	AP_{l}$\\
			\hline
			
			YOLOv8&\multirow{2}*{YOLO Series}&0.547&	0.945&0.582&0.271&0.552\\
			TSdetector&&0.575 \tiny +2.8\%&	0.955  \tiny+1.0\%&	0.625 \tiny+4.3\%&	0.353 \tiny+8.2\%&	0.577 \tiny+2.5\%\\
			\hline
			FCOS&\multirow{2}*{Image-Based}&0.475&	0.926&	0.423&	0.166&	0.479\\
			TSdetector&&0.512 \tiny+3.7\%&	0.944 \tiny+1.8\%&	0.482 \tiny+5.9\%&0.229 \tiny+6.3\%&	0.524 \tiny+4.5\%\\
			\hline
			RDN&\multirow{2}*{Video-Based}&0.437&	0.894&	0.371&	0.051&	0.440\\
			TSdetector&&0.452 \tiny+1.5\%&	0.901 \tiny+0.7\%&	0.399 \tiny+2.8\%&	0.155 \tiny+10.4\%&	0.461 \tiny+2.1\%\\
			\hline
			YOLOX&\multirow{2}*{YOLO Series}&0.524&	0.937&	0.545&	0.339&	0.512\\
			TSdetector&&0.564 \tiny+4.0\%&	0.953 \tiny+1.6\%&	0.612 \tiny+6.7\%&	0.384 \tiny+4.5\%&	0.566 \tiny+5.4\%\\
			\hline
	\end{tabular}}
	\label{backbone}
\end{table}

\begin{table}[!t] 
	\scriptsize  %%设置字体大小
	\setlength{\belowcaptionskip}{5pt}
	\caption{The ablation study of FPS and parameter quantities verified the impact of the proposed module on the model's computational complexity. GT-Conv and HQIM represent global temporal-aware convolution and hierarchical queue integration mechanism, respectively.}
	\centering
	\renewcommand\arraystretch{1.4}
	\resizebox{0.35\textwidth}{!}{
		\begin{tabular}{cc|cc}
			\hline
			GT-Conv&HQIM&FPS (f/s)&Params (M)\\
			\hline
			&&33.61&92.07\\
			\checkmark&&32.15 \tiny-1.46&94.11 \tiny+2.04\\
			&\checkmark&29.57 \tiny-4.04&96.57 \tiny+4.50\\
			\checkmark&\checkmark&28.29 \tiny-5.32&98.61 \tiny+6.54\\
			\hline
	\end{tabular}}
	\label{GT-ConvHQIM}
\end{table}

\textbf{Ablation for different backbone architectures.}  A broader perspective is provided by applying the proposed concepts to different backbone architectures. Given that our focus lies on CNN-based one-stage detection methods in this work, the backbone networks chosen are YOLOv8 for the YOLO series, FCOS for image-based, and RDN for the video-based method. The results demonstrate the positive impact of the proposed modules, albeit with varying performance enhancements across different backbone networks (Table \ref{backbone}). For the current state-of-the-art YOLO series network YOLOv8, $AP_{0.75}$ increased by 2.8\%. Notably, the most significant improvement is observed in $AP_{m}$, with an increase of 8.2\%, indicating that the module effectively enhances the polyp localization ability, making up the original deficiencies of the backbone network. Notably, the final results exceed those based on the YOLOX network in the original manuscript, indicating that choosing a better backbone network is more beneficial to the final performance. Additionally, notable improvements are observed in the FCOS backbone network, with increases of 3.7\%, 1.8\%, 5.9\%, 6.3\%, and 4.5\% in $AP_{0.5-0.75}$, $AP_{0.5}$, $AP_{0.75}$, $	AP_{m}$, $AP_{l}$, respectively.

\begin{figure}[!t]
	\centering
	\includegraphics[width=0.7\textwidth]{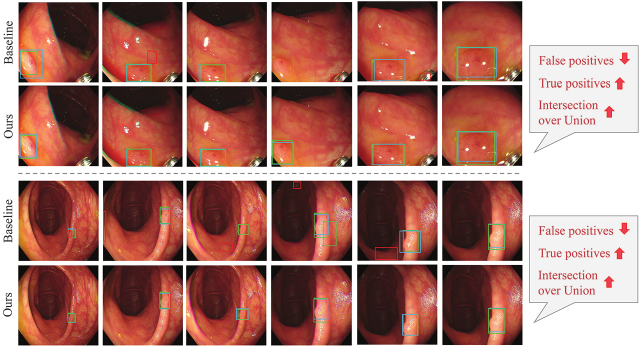}
	\caption{The visualization compared to the baseline method shows that TSdetector localizes more accurately, effectively reducing false positives while increasing true positives. Among them, the green, blue, and red boxes represent ground truth, true positives, and false positives, respectively.}
	\label{img10}
\end{figure}

\textbf{Ablation of GT-Conv and HQIM modules in computational complexity.} Introducing GT-Conv led to a decrease in FPS by 1.46 relative to the baseline, accompanied by an increase of 2.04 million parameters Table \ref{GT-ConvHQIM}. Conversely, HQIM introduces even greater computational overhead, resulting in a 4.04 FPS reduction and an increase of 4.50 million parameters. This finding indicates that the enhanced memory and aggregation capabilities offered by HQIM come at the expense of increased computational complexity. While the combination of GT-Conv and HQIM does elevate the computational complexity of our model, the resulting improvements in feature representation and object detection accuracy justify these additions.

\textbf{Visualization of TSdetector v.s. baseline model.} By comparing the visualization of the prediction results to the baseline, it can be clearly seen that TSdetector has a lower missed detection rate for continuous videos (Fig. \ref{img10}). Base detectors often struggle with false positives in bounding box prediction, leading to missed detections of polyps during ongoing lesion tracking. In contrast, TSdetector consistently delivers accurate predictions, substantially reducing the incidence of false positives. This improvement, in turn, enhances the alignment between the predicted bounding boxes and the actual ground truth.

\begin{figure}[!t]
	\centering
	\includegraphics[width=0.7\textwidth]{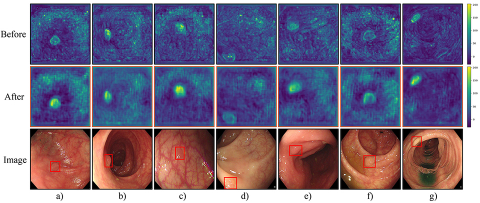}
	\caption{The visualization shows that the model pays more attention to the lesion area by comparing the feature maps before and after the hierarchical queue integration mechanism. The red box represents the ground truth.}
	\label{HQIM}
\end{figure}

\textbf{Effectiveness of HQIM module.} To verify the enhanced representation of features by the HQIM module, the feature maps before and after integrating the module are visualized in Fig. \ref{HQIM}. The experimental findings demonstrate that HQIM indeed effectively enhances the feature maps, thereby leading to an overall improvement in detection accuracy. Initially, without the HQIM module, feature maps exhibited limitations in accurately distinguishing target objects in complex backgrounds. However, upon applying HQIM, discernible enhancements are observed in feature map discriminability, with attention being directed away from irrelevant background elements and more focused on the target object. This improvement in feature representation can be attributed to HQIM 's ability to retain and propagate previous information to the current frame, thus facilitating the continuous integration of features across frames. By leveraging memory networks and employing a hierarchical propagation strategy, HQIM effectively captures rich features between frames, ensuring comprehensive processing of dynamic and temporal relationships within the data.

\begin{figure}[!t]
	\centering
	\includegraphics[width=\textwidth]{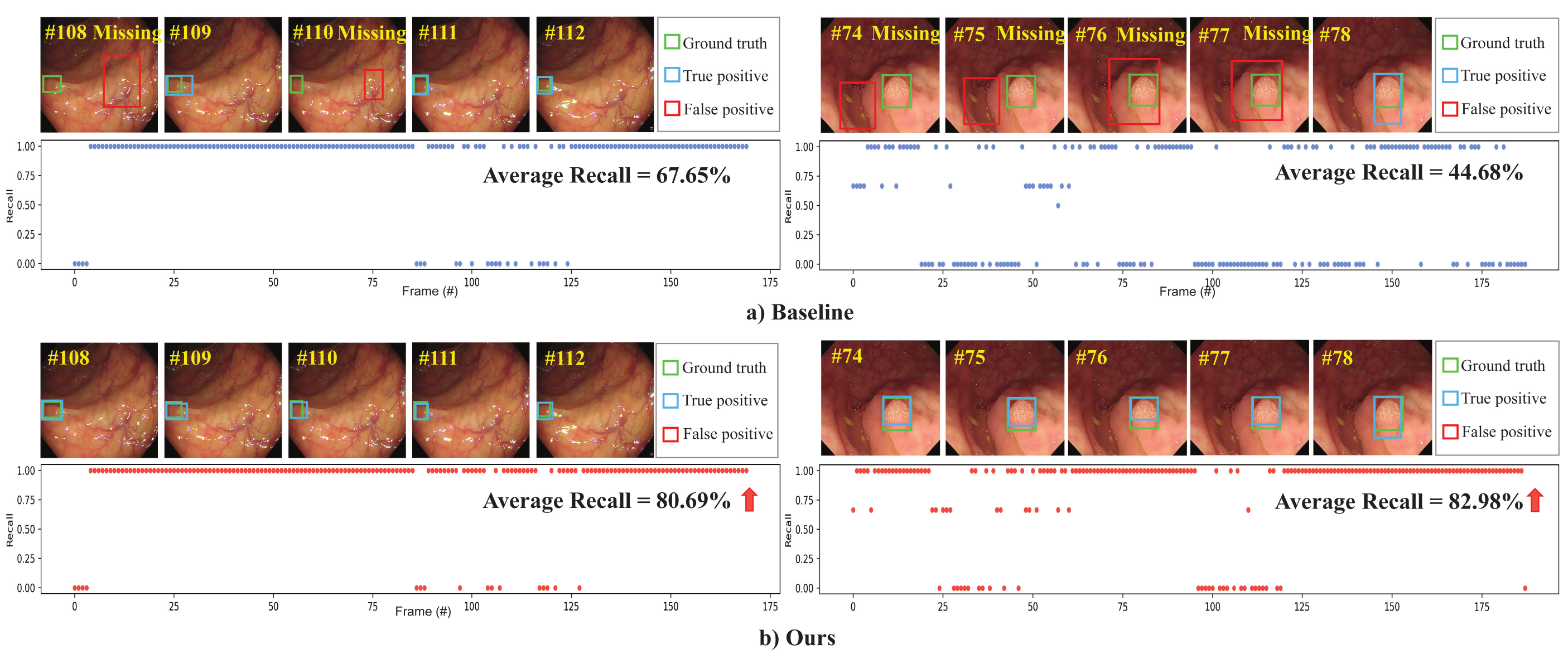}
	\caption{The frame-by-frame analysis of polyp video tracking shows that TSdetector can effectively improve the continuity of tracking, thereby avoiding the omission of targets. The upper and lower scatter plots represent the baseline and TSdetector, respectively.}
	\label{img11}
\end{figure}

\begin{figure}[!t]
	\centering
	\includegraphics[width=0.7\textwidth]{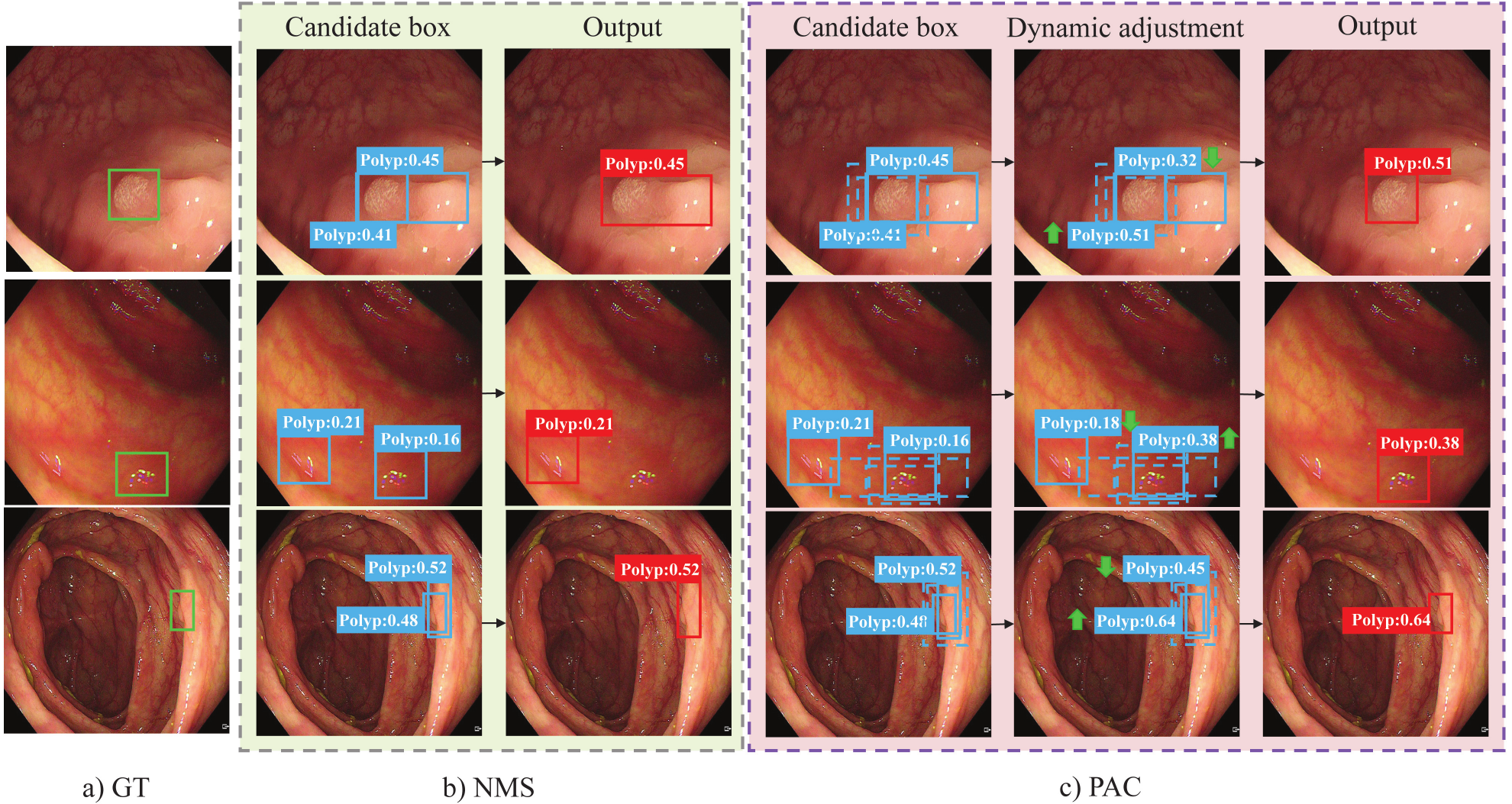}
	\caption{Examples where PAC can simultaneously enhance true positives and remove redundant boxes vs. classical NMS.}
	\label{img12}
\end{figure}

\subsection{Effectiveness Analysis}
\textbf{Analysis of the detection continuity.} To analyze the model's ability to cope with intra-sequence distribution heterogeneity during continuous localization of lesions, we evaluated its performance using a recall metric that quantifies the correct identification of positive samples in prediction results. A recall rate of R=100\% indicates an absence of missed detections. Fig. \ref{img11} visually represents our frame-by-frame trace recording. Our approach showcases remarkable stability and robustness (second row), capitalizing on temporal insights in contrast to the baseline detectors (first row). This can be attributed to the TSdetector's ability to model long-term dependencies in time and capture consistent features across frames. On the one hand, convolution kernels, informed by prior knowledge, dynamically guide the extraction of consistent features. On the other hand, the model possesses long memory capabilities to facilitate hierarchical inter-frame feature aggregation. Consequently, our detectors consistently demonstrate reliable performance even within the intricate array of endoscopic tracking conditions.

\textbf{Analysis of the adaptive confidence.} To assess the impact of the post-processing module on the difference between confidence and accuracy in the model, we conducted a comparative analysis of two post-processing methods: traditional NMS and PAC. In Fig. \ref{img12}, illustrates how adaptive confidence, as implemented by PAC, enhances the accuracy of polyp localization.  PAC showcases the capability to not only reduce confidence scores associated with false negative boxes but also increase the confidence values linked to true positive boxes, as compared to conventional methods. This dual effect allows it to effectively distinguish false detections from valid ones while improving the overall confidence estimate. The adjustment of post-confidence for candidate frames based on their positional relationships prevents the erroneous exclusion of highly accurate frames due to confidence considerations, resulting in consistently superior results.

\begin{figure}[!t]
	\centering
	\includegraphics[width=0.6\textwidth]{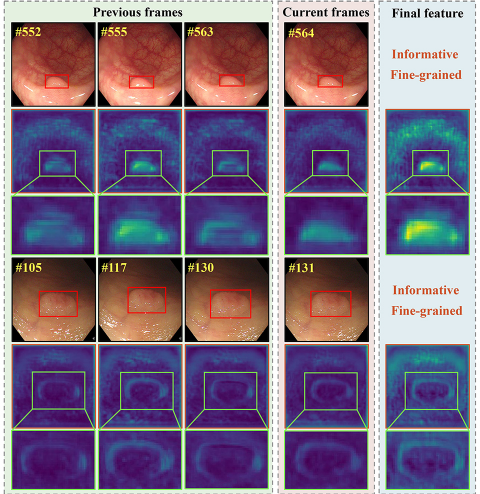}
	\caption{The comparison visualization shows the informative and fine-grained features obtained, including the current frame, previous frames, and aggregated features. The red box represents the ground truth; the green box represents the feature map near the polyp.}
	\label{informative}
\end{figure}

\textbf{Effectiveness of aggregation of multi-temporal features.} The results show that informative and fine-grained features can be obtained as described by visualizing the features of the current frame and multiple previous frames and comparing them with the integrated features. As shown in Fig. \ref{informative}, it can be seen that the feature map of a single frame always only pays attention to the polyp part, often failing to encompass all aspects of the polyp and lacking clear contrast with irrelevant background areas. Conversely, aggregated features focus on polyp details and emphasize polyp edges and textures, enhancing differentiation from surrounding tissues.

\begin{figure}[!t]
	\centering
	\includegraphics[width=0.7\textwidth]{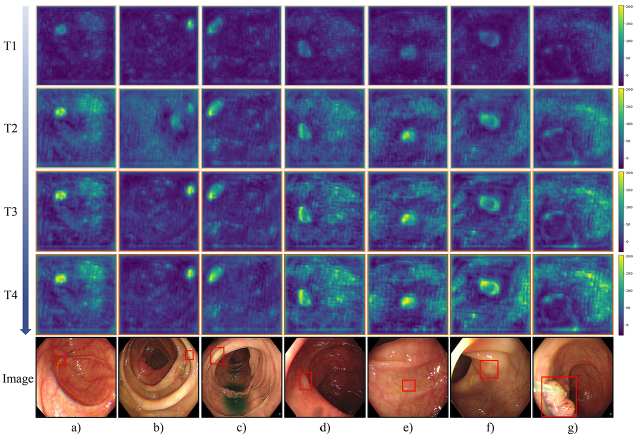}
	\caption{The visualization of features at each layer of the progressive accumulation mechanism shows the process of knowledge accumulation between frames. The red box represents the ground truth.}
	\label{accumulation}
\end{figure}

\textbf{Effectiveness of temporal knowledge aggregation.} The feature maps in the progressive accumulation mechanism (Fig. \ref{accumulation}) clearly show the progressive evolution of multi-temporal features. There is a discernible intensification in focus on the target area, accompanied by an increasingly pronounced differentiation from the background. Additionally, the detected targets become more evident over time. These observations prove that our method effectively exploits the extracted temporal context information, improving inter-frame consistency. Through dynamic knowledge aggregation across frames, our approach reinforces feature representations, fostering a more robust and coherent understanding of temporal dynamics.

\begin{table}[!t] 
	\scriptsize  %%设置字体大小
	\setlength{\belowcaptionskip}{5pt}
	\caption{The proposed method has a lower false positive rate compared to the baseline model when tested on negative videos on the SUN dataset.}
	\centering
	\renewcommand\arraystretch{1.5}
	\resizebox{0.7\textwidth}{!}{
		\begin{tabular}{c|c|c|c}
			\hline
			Method&	Number of negative frames&	Number of false positive boxes detected&	False positive rate\\
			\hline
			YOLOX&	\multirow{2}*{109554}	&18871&	17.23\%\\
			TSdetector&&11180 \tiny-7691&	10.21\% \tiny-7.02\%\\
			\hline
	\end{tabular}}
	\label{negative}
\end{table}

\begin{figure}[!t]
	\centering
	\includegraphics[width=0.7\textwidth]{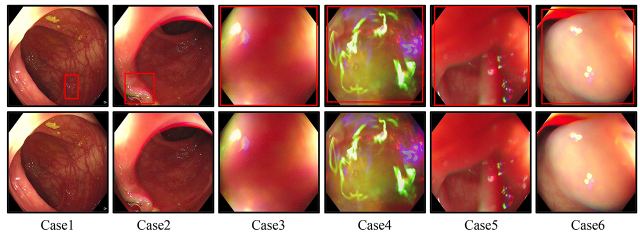}
	\caption{Visualization example on negative video compared to the baseline model.}
	\label{negativepdf}
\end{figure}

\textbf{Effectiveness on negative videos.} TSdetector is tested on negative videos and compared with baseline models, validating its performance in reducing false positives (Table \ref{negative}). Specifically, the SUN dataset comprises 13 negative videos and a total of 109554 frames of images. The baseline model exhibited a false positive rate of 17.23\%, resulting in 18871 false positive boxes. In contrast, TSdetector significantly reduced false positives, with only a 10.21\% false positive rate and 7691 false positive boxes. Visual results (Fig. \ref{negativepdf}) illustrate that TSdetector utilizes the correlation between nearby frames to effectively weaken the bias observed in a single image, thereby improving the reliability of the model.

\begin{figure}[!t]
	\centering
	\includegraphics[width=\textwidth]{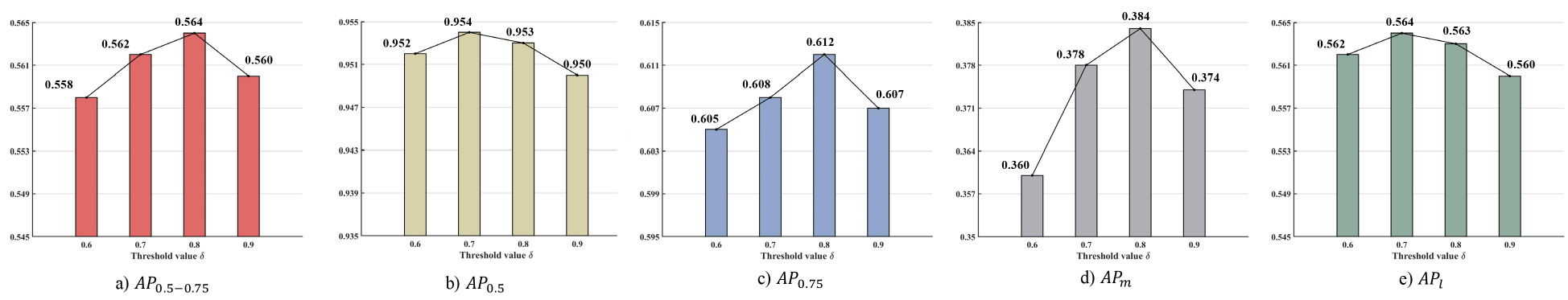}
	\caption{Quantitative results of the impact of threshold $\delta$ parameters on the model in the SUN dataset, including metrics: a) $AP_{0.5-0.75}$, b) $AP_{0.5}$, c) $AP_{0.75}$, d) $	AP_{m}$, e) $	AP_{l}$.}
	\label{delta}
\end{figure}

	\subsection{Hyperparameter Analysis.} \textbf{Quantitative impact of threshold $\delta$.} The threshold $\delta$ has little impact on the overall performance, and the fluctuation of $AP_{0.5-0.75}$ is within 0.6\%. The threshold $\delta$ affects the number of boxes included in the friend box set, meaning that the larger the threshold, the fewer friends the current box has. Fig. \ref{delta} illustrates the results of $AP_{0.5-0.75}$,	$AP_{0.5}$,	$AP_{0.75}$, $	AP_{m}$, $	AP_{l}$ under different $\delta$ values of 0.6, 0.7, 0.8, and 0.9 respectively. Overall, the model achieves optimal performance when $\delta$ is around 0.8. As the $\delta$ increases, fewer friend boxes remain, leading to higher overlap rates among them. Consequently, it can be seen the detection rate of small polyps $AP_{m}$ is increased, and $AP_{0.5}$ is higher, indicating that the positioning accuracy is improved. Moreover, changes in the parameter have less impact on $AP_{l}$ (large polyps) and $AP_{0.5}$. However, an excessively large threshold adversely affects overall performance, indicating that reducing the number of friend boxes diminishes the algorithm's effectiveness.

\subsection{Model Limitations and Future Directions.} 
	An observation emerged by visualizing the prediction results for negative videos: many false positives arise from image corruption, such as water washout and probe adhesion during colonoscopy. This is because the model training data is entirely derived from annotated partial colonoscopy videos containing lesions and, therefore, mainly contains clear images. However, during testing, the negative sample data contained a complete colonoscopy video depicting all the complications encountered during the procedure, as shown in Fig. \ref{negativepdf}. To address this issue, future efforts will focus on two key strategies: 1) Pre-discrimination of data: The model can be configured to ignore damaged frames that do not require prediction, thereby reducing the impact of image corruption on the overall performance. 2) Integration of unsupervised learning mechanisms: Incorporating an unsupervised learning framework will enable the model to learn unlabeled negative samples, enhancing its adaptability in real-world scenarios.

\section{Conclusion}
\label{sec6}
This paper introduces a novel framework for a temporal-spatial self-correcting detector, which highlights how collaborative learning can be used to utilize address critical challenges in the field of video polyp detection. To the best of our knowledge, this is the first trial exploring a one-stage architecture based on temporal- and spatial-level optimization for the continuous detection of polyp lesions. We first build a global temporal-aware convolution to adjust the convolution kernel, enabling feature calibration through contextual information. Then, a hierarchical queue integration mechanism is designed to endow the model with long-term memory capabilities, facilitating information propagation within time series data. Finally, position-aware clustering is employed to further dynamically correct the confidence score. The results demonstrate that TSdetector achieves a polyp detection rate of up to 95.30\%, outperforming the baseline and seventeen state-of-the-art methods. We assert that TSdetector holds the potential to serve as a powerful and reliable tool for real-time polyp detection, further advancing the development of colonoscopy.

\section*{Acknowledgments}
This work was supported National Key Research and Development Program of China (2018YFA0704102), in part by the National Natural Science Foundation of China (62371121), and in part by the Jiangsu Provincial Key R\&D Program, China (BE2022827).

%%Harvard
\bibliographystyle{model2-names.bst}\biboptions{authoryear}
\bibliography{refs}

%\section*{Supplementary Material}

%Supplementary material that may be helpful in the review process should
%be prepared and provided as a separate electronic file. That file can
%then be transformed into PDF format and submitted along with the
%manuscript and graphic files to the appropriate editorial office.

\end{document}